\documentclass[lettersize,journal]{IEEEtran}
\usepackage{amsmath,amsfonts}
\usepackage{algorithmic}
\usepackage{algorithm}
\usepackage{array}
\usepackage[caption=false,font=normalsize,labelfont=sf,textfont=sf]{subfig}
\usepackage{textcomp}
\usepackage{stfloats}
\usepackage{url}
\usepackage{verbatim}
\usepackage{hyperref}
\usepackage{graphicx}
\usepackage{cite}
\usepackage{multirow}
\usepackage[table]{xcolor}
\usepackage{makecell}

\begin{document}

\title{PhysiInter: Integrating Physical Mapping for High-Fidelity Human Interaction Generation}

\author{Wei~Yao,
        Yunlian~Sun,
        Chang~Liu,
        Hongwen~Zhang,
        and~Jinhui~Tang
\thanks{Wei Yao, Yunlian Sun and Jinhui Tang are with the School of Computer Science and Engineering, Nanjing University of Science and Technology,
Nanjing 210094, China. (e-mail: wei.yao@njust.edu.cn; yunlian.sun@njust.edu.cn; jinhuitang@njust.edu.cn)}
\thanks{Chang Liu and Hongwen Zhang are with the School of Artificial Intelligence, Beijing Normal University, Beijing 100875, China. (email: lc826@mail.bnu.edu.cn, zhanghongwen@bnu.edu.cn)}
}

\markboth{Journal of \LaTeX\ Class Files,~Vol.~14, No.~8, August~2021}%
{Shell \MakeLowercase{\textit{et al.}}: A Sample Article Using IEEEtran.cls for IEEE Journals}


\maketitle

\begin{abstract}
Driven by advancements in motion capture and generative artificial intelligence, leveraging large-scale MoCap datasets to train generative models for synthesizing diverse, realistic human motions has become a promising research direction. However, existing motion-capture techniques and generative models often neglect physical constraints, leading to artifacts such as interpenetration, sliding, and floating. These issues are exacerbated in multi-person motion generation, where complex interactions are involved. To address these limitations, we introduce physical mapping, integrated throughout the human interaction generation pipeline. Specifically, motion imitation within a physics-based simulation environment is used to project target motions into a physically valid space. The resulting motions are adjusted to adhere to real-world physics constraints while retaining their original semantic meaning. This mapping not only improves MoCap data quality but also directly informs post-processing of generated motions. Given the unique interactivity of multi-person scenarios, we propose a tailored motion representation framework. Motion Consistency (MC) and Marker-based Interaction (MI) loss functions are introduced to improve model performance. Experiments show our method achieves impressive results in generated human motion quality, with a 3\%–89\% improvement in physical fidelity. Project page \href{https://yw0208.github.io/physiinter/}{\textcolor{magenta}{https://yw0208.github.io/physiinter/}}.
\end{abstract}


\section{Introduction}
\IEEEPARstart{I}{n} embodied artificial intelligence, synthesizing high-fidelity human motions is a crucial challenge with broad applications, such as animation, gaming, virtual reality, augmented reality, and intelligent robotics. Text serves as the most fundamental communication medium. Consequently, text-to-motion generation has become one of the most promising generative AI tasks. Advances in motion capture (MoCap) technology~\cite{pymaf, pymaf-x, yao2024staf, vibe, spec} have enabled efficient acquisition of large-scale human motion datasets. The newly emerged diffusion model~\cite{ho2020denoising}, leveraging its robust capacity for modeling complex distributions, excels in generating intricate motion sequences. Driven by extensive datasets and innovative model architectures, numerous text-to-motion approaches~\cite{tevet2023mdm, shafir2023priormdm, zhang2024motiondiffuse, Zhang2023t2mgpt, Jiang2023MotionGPTHM, Tevet2022MotionCLIPEH, Zhang2024motionmamba, Lu2023HumanTOMATOTW, Liang2023OMGTO} have been proposed in recent years.

\begin{figure}[ht]
    \centering
    \includegraphics[width=\linewidth]{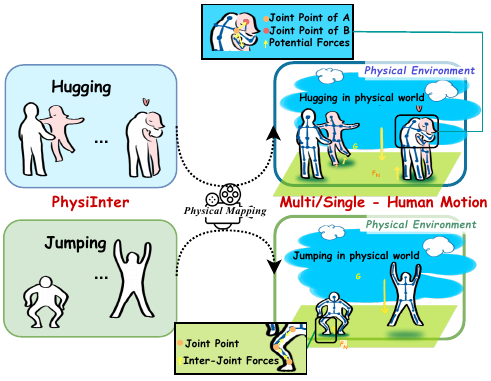}
    \caption{\textbf{Insight} All human activities inherently adhere to the laws of physics. Motivated by this observation, we introduce physical mapping, a framework that enforces physical constraints on human motion generation, thereby producing higher-fidelity motion sequences.}
    \label{fig: first}
\end{figure}

As illustrated in Fig.~\ref{fig: first}, human motion generation poses significant challenges due to complex human interactions. Both MoCap systems and generative models often overlook the fact that human motion adheres to physical laws. Compared to single-person MoCap, multi-person scenarios introduce challenges such as occlusion, contact, and relative positioning, which degrade data quality. If training data fails to adhere to physical constraints, the learned motion distribution will deviate from physical realism. Even with high-quality training data, generative models still cannot ensure physically valid results. Diffusion models, for instance, primarily rely on statistical fitting of motion data distributions and lack explicit physical constraints. This deficiency leads to artifacts such as foot sliding, floating, ground penetration, and interpenetration between humans. Thus, endowing models with physical awareness is critical to addressing these issues.

\begin{figure*}[ht]
    \centering
    \includegraphics[width=0.99\textwidth]{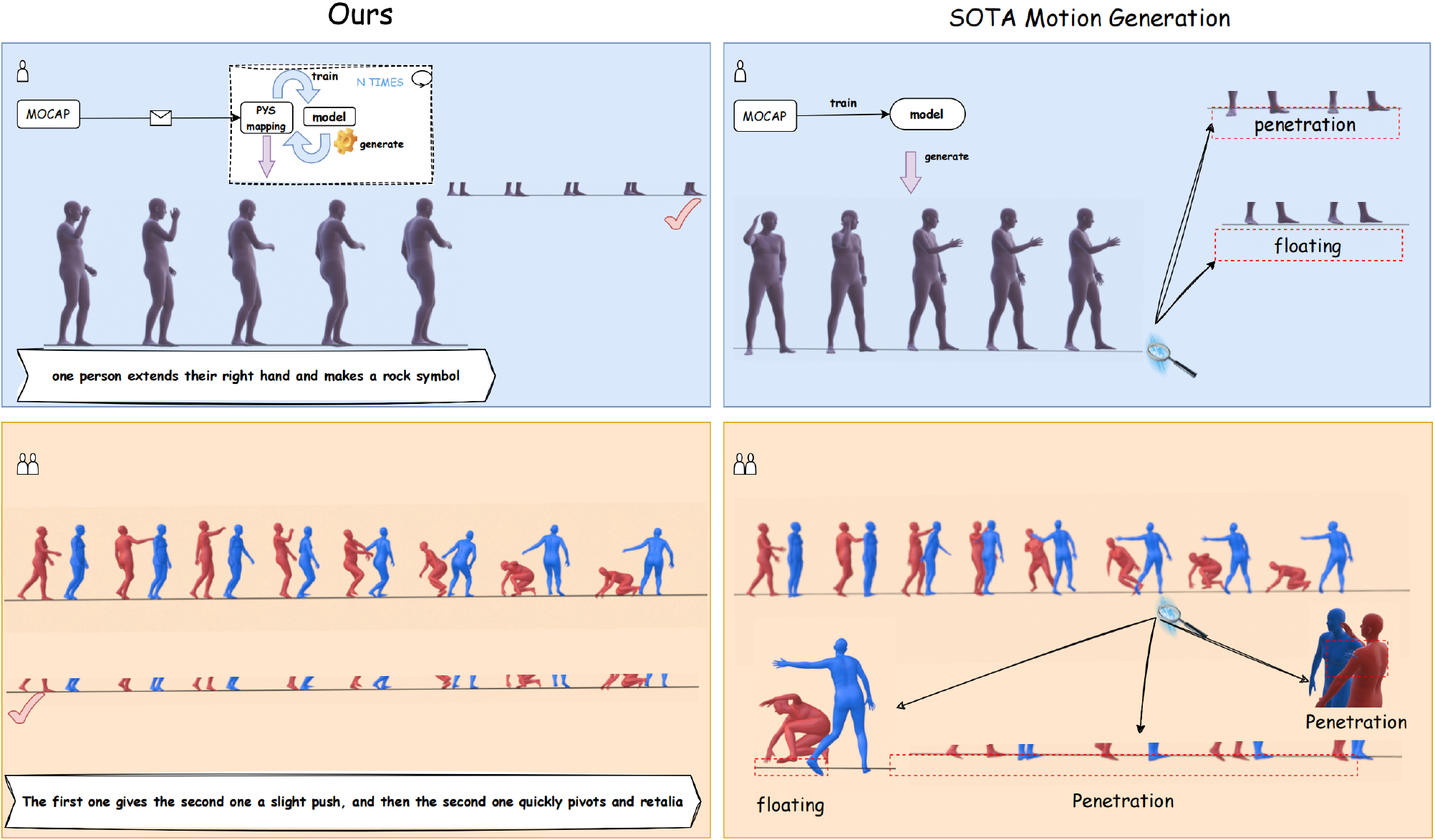}
    \caption{\textbf{Comparison with Previous Methods} Prior approaches directly leverage \textbf{mo}tion \textbf{cap}ture data to train generative models, which often results in motion artifacts. To address this, we propose physical mapping, a method that preprocesses mocap data during training and directly optimizes the generated motion sequences, thereby significantly enhancing motion quality.}
    \label{fig: first}
\end{figure*}

To address these challenges, we introduce \textbf{physical mapping}, a method that integrates physical constraints into both training data and generative models. Directly modeling real-world physics is intractable. While physical simulation tools~\cite{makoviychuk2021isaac, todorov2012mujoco} can approximate real-world dynamics, they are difficult to combine with generative models directly. To bridge this gap, we employ motion imitation, where agents in a physics-based simulation environment replicate input motions. Specifically, we create one or more humanoid agents with human-like physical properties (e.g., mass, inertia, collision detection). These agents are trained to mimic target motions while adhering to physical laws. The resulting motions retain the semantic intent of the original motion while satisfying physical constraints. We term this process physical mapping. By enforcing physics constraints on both training data and generated outputs, our approach significantly enhances motion quality.

Multi-person interaction generation~\cite{Ghosh2023REMOS3M, liang2024intergen, xu2024interx, ruiz2024in2in, Javed2024InterMask3H, Li2024InterDanceReactive3D, Huang2024InterActCA, Xu2024ReGenNetTH, Tanaka2023RoleawareIG} differs fundamentally from single-person motion generation, as it requires modeling interpersonal interactions. Since relative positions between individuals must be generated, the canonical skeletal representation used in prior work~\cite{guo2022generating} is insufficient. To address this, we introduce a non-canonical motion representation that encodes complete global motion data. Differing from \cite{liang2024intergen}, we add global orientation and remove redundant foot contact labels to facilitate model learning and deployment. Moreover, existing models underutilize the rich information in motion representations during training and inference. To mitigate this, we propose two novel loss functions: Motion Consistency (MC) loss and Marker-Based Interaction (MI) loss.
MC loss ensures consistency across motion components (position, velocity, rotation) during diffusion-based generation. This yields three key benefits: (1) It maintains plausible human bone lengths by enforcing kinematic constraints.
(2) It provides stronger supervision signals for model training. (3) It improves rotation quality, as joint rotations are critical in practical applications (e.g., animation).
MI loss shifts the focus from joint positions to sparse surface markers on human meshes. Since human interaction primarily involves surface contact, markers better capture interaction semantics than joints or vertices. This enables more precise and efficient modeling of interpersonal dynamics.

In summary, our contributions are as follows:

\begin{itemize}
    \item[$\bullet$]We introduce physical mapping, a two-pronged approach to integrat physics into text-to-motion generation. Physical mapping refines raw MoCap data to align with physical constraints, improving training data quality for generative models. And it directly constrains generative models to produce physically valid motions during inference.
    \item[$\bullet$]We enhance model capacity for multi-person interaction learning through: (1) A non-canonical human interaction representation that encodes global motion semantics. (2) Novel MC and MI loss that enforce motion consistency and interaction precision.
    \item[$\bullet$]Experiments on the large-scale motion dataset~\cite{liang2024intergen, guo2022generating} demonstrate significant improvements in dataset quality and generation quality, particularly in physical fidelity.
\end{itemize}

\section{Related Work}

\subsection{Text-Driven Human Motion Generation}
\textbf{Single Human Motion Generation}
In recent years, generative models have seen rapid advancements, with motion generation emerging as a key focus in embodied artificial intelligence. Early works~\cite{huang2020dance, ghosh2021synthesis, plappert2018learning} modeled motions directly and established feature mappings conditioned on inputs but produced stereotyped, average motions. The introduction of generative adversarial networks (GANs)~\cite{goodfellow2014gan}, variational autoencoders (VAEs)~\cite{Lopez2020vae}, and diffusion models~\cite{Ho2020diffusion} has spurred a wave of innovations. Currently, VAEs and diffusion models, along with their variants, dominate the field. Notable VAE-based works include TEMOS~\cite{petrovich2022temos}, T2M-GPT~\cite{Zhang2023t2mgpt} and MLD~\cite{chen2023mld}, which used a Transformer-based VAE to bridge language and motion spaces. And MoMask~\cite{guo2024momask} combined residual vector quantization (RVQ-VAE~\cite{Lai2022RVQ}), masked transformers, and residual transformers to achieve SOTA performance. Diffusion models, such as MDM~\cite{tevet2023mdm}, MotionDiffuse~\cite{zhang2024motiondiffuse}, PhysDiff~\cite{yuan2023physdiff}, PriorMDM~\cite{shafir2023priormdm}, and DiffusionPhase~\cite{wan2023diffusionphase}, have demonstrated superior data modeling capacity, enabling diverse realistic human motion synthesis.

\textbf{Multi-Person Interaction Generation}
Human motion is inherently interactive. While single-person generation has developed a lot, multi-person interaction remains underexplored yet critical for real-world applications. Early works like ComMDM~\cite{shafir2023priormdm} extended single-person models by combining two frozen MDMs~\cite{tevet2023mdm} with a cross-attention module. Subsequent works~\cite{liang2024intergen, ruiz2024in2in, Javed2024InterMask3H} adopted similar cross-attention architectures. DLP~\cite{cai2024digital} and Sitcom~\cite{Chen2024SitcomCrafterAP} further integrated social intelligence, enabling agents to communicate and do motions, advancing beyond basic text-to-motion generation.

\textbf{Limitations}
Despite progress, existing methods overlook physical constraints due to data and algorithmic limitations. Human motion datasets, e.g., AMASS~\cite{mahmood2019amass} for single-person data, InterHuman~\cite{liang2024intergen}, and Inter-X~\cite{xu2024interx} for multi-person data, lack physical fidelity. Motion capture inaccuracies omit critical interactions like collisions and friction, which leads to artifacts like sliding, penetration, and floating. Generative models exacerbate these issues by learning motion distributions without physical grounding. While post-processing can mitigate artifacts, it often sacrifices motion naturalness, failing to balance physical plausibility and realism.

\subsection{Physics-based Motion Imitation}
\textbf{Definition} In physics-based simulation environments, simulated agents’ behaviors are governed by physical constraints. This is particularly advantageous for tasks involving human-human or human-object interactions. Physics-based motion imitation involves training simulated agents to replicate input motion sequences within a physically constrained environment, actually serving as a form of motion retargeting. Imitating complex human motions has long been a foundational challenge. Early works like RFC~\cite{yuan2020rfc} used residual forces to stabilize agents during imitation. Later, adversarial reinforcement learning-based methods \cite{peng2022ase, peng2021amp} expanded agents’ motion skills. State-of-the-art methods such as UHC~\cite{luo2023uhc} and PHC~\cite{luo2023phc} achieve near 100\% imitation success on AMASS~\cite{mahmood2019amass} via expert policy mixtures.

\textbf{Application} Motion imitation has two primary applications. First, motion imitation can be used for skill acquisition, i.e., training low-level policies to enable agents to perform complex tasks. Works like PhysHOI~\cite{wang2023physhoi}, AnySkill~\cite{cui2024anyskill}, and H. Zhang et al.~\cite{yuan2023tennis} used imitation to teach agents skills such as playing basketball, door-knocking, and playing tennis.
Second, motion imitation can enhance pose estimation. Addressing issues like jitter, occlusion, and penetration in human mesh recovery. Methods like SimPoE~\cite{yuan2021simpoe}, Embodied-Pose~\cite{luo2022embodiedscene}, and MultiPhys~\cite{ugrinovic2024multiphys} leveraged imitation for robust pose estimation. A unique example is Ego-Pose~\cite{yuan2019ego}, which uses PD-control-based imitation to predict 3D poses from egocentric videos.
As for us, we propose a third application integrating motion imitation into the full motion generation pipeline to enhance interaction realism.

\begin{figure*}[ht]
    \centering
    \includegraphics[width=0.99\textwidth]{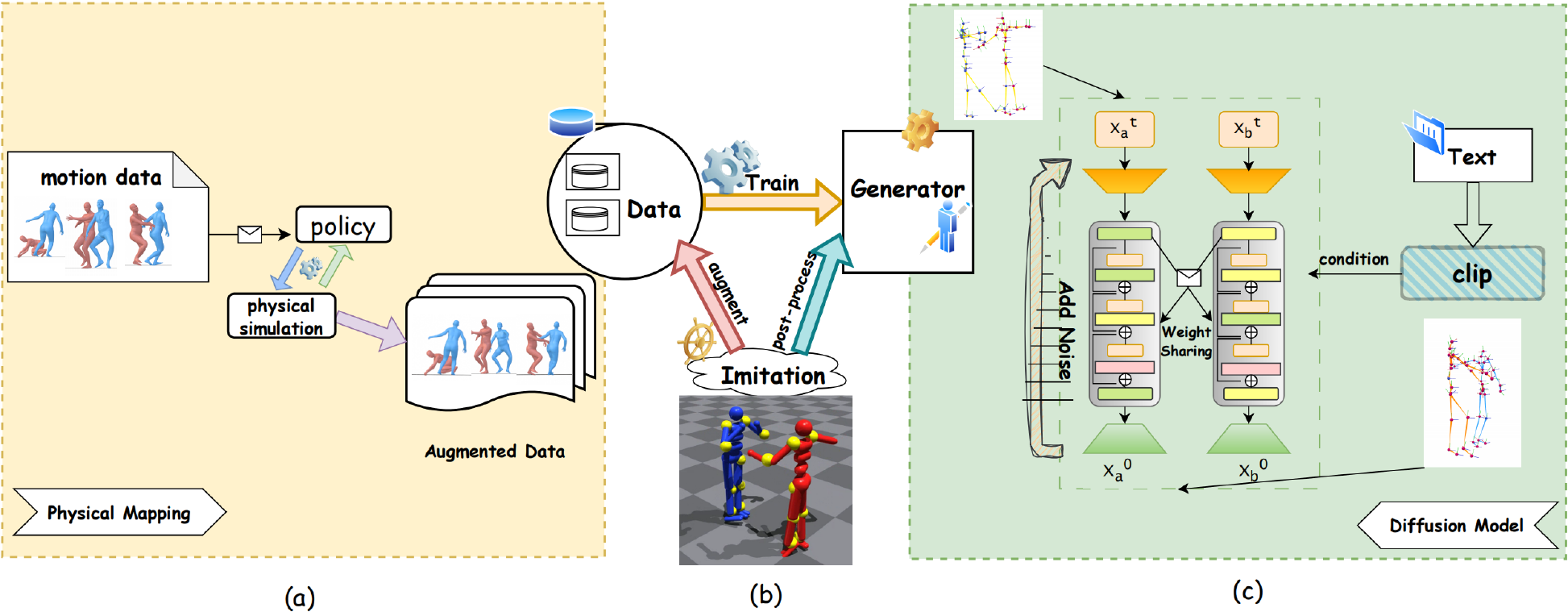}
    \caption{
    \textbf{Overview of Pipeline}
    (a) Data: A reinforcement learning policy is employed to map motion data into a physical simulator. This step enhances motion data quality while ensuring physical plausibility.
    (b) Imitation: Motion imitation serves as the core component of our framework, enabling data augmentation and post-processing of generated motions.
    (c) Generator: A diffusion model is utilized, guided by text prompts to denoise motion sequences, thereby enabling text-to-motion synthesis. Note that our framework does not impose constraints on the generator architecture; the diffusion model is provided as an example.
}
    \label{fig: model}
\end{figure*}

\section{Method}

This section outlines our three core contributions, each detailed in subsequent subsections. In Section~\ref{sec: motion_rep}, we introduce a motion representation tailored for multi-person interactions and compare it with prior approaches. Section~\ref{sec: phys_map} presents the overall pipeline of our method, including a detailed explanation of physical mapping and its implementation. Section~\ref{sec: loss} describes our loss functions, with a focus on the newly proposed motion consistency loss and marker-based interaction loss. For clarity, we use two-person motion generation as an example throughout this section. Note our method generalizes to any number of participants, and its applicability to single-person scenarios is validated in the experimental results.  

\subsection{Motion Representation}\label{sec: motion_rep}

\textbf{Preliminary} Early methods predominantly relied on joint position coordinates for human motion characterization. With the prevalence of parametric human mesh models such as SMPL and its extended version SMPL-X, contemporary approaches~\cite{liu2023emage, yi2023talkshow, qi2024cocogesture} have adopted pose parameters as primary motion descriptors, particularly in speech-driven gesture generation where spatial movement consideration is unnecessary. For text-to-motion generation, the pioneering work of Guo et al.~\cite{guo2022generating} introduced a canonical motion representation comprising seven components: root angular velocity, root linear velocity, root height, joint positions, joint velocities, global joint rotations (excluding root joint), and foot contact labels. This canonical coordinate system constrains motion initiation to the origin $(0, 0)$, making it unsuitable for multi-person interaction requiring global positional control. 

Note that the terms \textbf{position-based} (-pos) and \textbf{rotation-based} (-rot) appear frequently in subsequent sections. Motion representations comprise both joint positions and rotations, where each representation can fully characterize motion. A \textbf{position-based} approach recalculates other motion components (e.g., rotations) from joint positions, whereas a \textbf{rotation-based} approach derives other components (e.g., positions) from joint rotations. We subsequently demonstrate that generative models struggle to maintain internal consistency between these motion representation components, which adversely affects both model training and inference.

\textbf{Human Interaction Representation} First, the task of multi-person motion generation needs to be defined. Given conditions $c$, i.e.,  text describing human motions, the generative model is expected to produce a sequence of human motion representations $X\in \mathbb{R} ^{N\times T\times L}$. Here, $N$ denotes the number of humans, $T$ represents a fixed length of human motion sequences, and $L$ indicates the dimension of motion representation. In this study, the motion representation $\left[ p, v, r \right]$ consists of three components: joint position, joint velocity, and joint rotation. Consistent with SMPL~\cite{loper2015smpl} and SMPL-X~\cite{smplx}, 22 SMPL joints (excluding hand joints) are utilized. The joint position is represented as a 3D global coordinate, while joint velocity is a 3D vector indicating the position transformation of the joint from the previous frame to the subsequent frame. Joint rotation is expressed as a local joint rotation using a 6D rotation matrix. Consequently, the motion representation has a length of $L=22\times3+22\times3+22\times6=264$. Once $X\in \mathbb{R} ^{N\times T\times L}$ is generated, users may opt to utilize the position component for skeleton-based animation or the rotation component to drive an SMPL mesh, depending on specific requirements.

\textbf{Comparison} Recent advancements by Liang et al.~\cite{liang2024intergen} proposed non-canonical representation using global root positions and local joint rotations. Our architecture extends this foundation by incorporating root joint rotation while eliminating foot contact labels, yielding three key improvements. First, the inclusion of root joint rotation enables direct implementation of rotational components, bypassing the computationally intensive mesh fitting procedures. Second, our motion representation permits fast integration of MC loss during training, as detailed in Section~\ref{sec: loss}. Third, the removal of foot contact labels eliminates redundant constraints. While previous methods attempted to leverage these labels for ground contact regularization, our empirical analysis reveals they introduced noise rather than beneficial guidance.

\subsection{Pipeline}\label{sec: phys_map} 

\textbf{Motion Imitation}
As the technical foundation of our method, we provide an introduction to our motion imitation, following a framework proposed by Luo et al.~\cite{luo2023phc}. As depicted in Figure.~\ref{fig: model} (a), motion imitation is fundamentally a reinforcement learning (RL) approach executed within a physics-based simulator. RL is characterized by four core components: states $\mathcal{S}$, actions $\mathcal{A}$, policies $\pi$, and rewards $\mathcal{R}$. We elaborate on motion imitation through these elements. In our task, the states $\mathcal{S}$ encompass goal-oriented states and humanoid proprioceptive inputs, such as the simulated agent’s joint angles, velocities, and positions. The policy $\pi \left( a^h|s^h \right)$ generates actions $a^h \in \mathcal{A}$ based on the current state $s^h \in \mathcal{S}$, which are then applied to the agent via proportional-derivative (PD) controllers. The physics simulator subsequently updates the agent’s state to produce the subsequent state $s^{h+1}$. To train the policy, a reward function inspired by prior work \cite{luo2023phc, peng2021amp, peng2018deepmimic, fu2023deepwholebody} is adopted, decomposing $\mathcal{R}$ into a task reward, a style reward, and an energy penalty. The task reward incentivizes the agent to match the target joint angles. The style reward ensures the agent’s motion distribution aligns with real human motion patterns. The energy penalty discourages excessive mechanical energy expenditure, thereby promoting energy-efficient actions and mitigating motion jitter. In summary, motion imitation operates as a Markov decision process. Given a target motion sequence $x_{1:T}$, the agent generates a physically plausible motion sequence $\hat{x}_{1:T}$ by iteratively sampling states and actions.

\textbf{Physical Mapping}
Motion imitation enables two key applications in our framework, as illustrated in Figure~\ref{fig: model} (b): data augmentation and post-processing. We first address the augmentation component. In our pipeline, the target motion sequence $p_{1:T}$ denotes 24 SMPL joint positions, including 2 extra hand joints. This sequence $p_{1:T}$ is equivalent to the motion representation ${X}_{1:T}$ proposed in this work. While physical mapping operates on $p_{1:T}$ to get $\hat{p}_{1:T}$. The $\hat{p}_{1:T}$ can be losslessly converted to ${X}_{1:T}$. And our generative model uses ${X}_{1:T}$ for training and inference.  We preprocess motion capture datasets into joint position format. The motion imitation policy, originally designed for single-agent scenarios, is fine-tuned to support parallel simulation of multiple agents within the same collision group. This allows physical mapping to model human-human interactions, enhancing collision plausibility without restrictions on the number of agents.
The process yields a physically plausible joint position sequence $\hat{p}_{1:T}$. After discarding failed imitation samples, $\hat{p}_{1:T}$ is transformed into our motion representation $\hat{X}_{1:T}$. Joint velocity is computed as $\hat{v}=\hat{p}_{2:T+1}-\hat{p}_{1:T}$, and joint rotation $\hat{r}_{1:T}$ is derived via iterative optimization using a fitting method~\cite{bogo2016keep}.

The second application of physical mapping is post-processing to refine generated outputs. Although the generative model is trained on augmented data, residual physical inconsistencies may persist due to the absence of explicit constraints. To address this, we extract the positional component of the generated sequence and apply physical mapping to produce the final result. This procedure is similar to the augmentation workflow described above.

\textbf{Motion Generator} The architecture of the motion generator is only secondary to its compatibility with the motion units used in physical mapping. Provided that the generator employs a data-driven approach and aligns with the motion units used in physical mapping, its performance can be enhanced via our physical mapping. But for clarity, we briefly describe the motion generator used in our framework. As depicted in Figure~\ref{fig: model} (c), for two-person motion generation, we employ a conditioned diffusion model. The model’s process is formalized as:
\begin{equation}
    \left\{ \begin{array}{c}
	X_{a}^{\left( 0 \right)}=D_{\theta}\left( X_{a}^{\left( t \right)}+\sigma _t\epsilon _a, X_{b}^{\left( t \right)}+\sigma _t\epsilon _b, t, c \right)\\
	X_{b}^{\left( 0 \right)}=D_{\theta}\left( X_{b}^{\left( t \right)}+\sigma _t\epsilon _b, X_{b}^{\left( t \right)}+\sigma _t\epsilon _b, t, c \right)\\
\end{array} \right. 
\end{equation}
where $D_{\theta}$ denotes the shared denoising model with parameters $\theta$, $\sigma$ is the diffusion coefficient controlling noise intensity, $\epsilon \in \mathcal{N} \left( 0, I \right)$ represents Gaussian noise, $t$ is the time step, and $c$ is the text condition encoded via CLIP~\cite{radford2021clip}. Here, $a$ and $b$ index distinct persons. Iterative denoising yields the clean motion sequences $\left\{ X_{a}^{\left( 0 \right)}, X_{b}^{\left( 0 \right)} \right\}$ as the final output. For single-person motion generation, we adopt MDM~\cite{tevet2023mdm}, which shares a similar architecture but is tailored for single-person scenarios.

\subsection{Losses}\label{sec: loss} 

\textbf{Conventional Losses} This section outlines the loss functions commonly employed in human motion generation. As depicted in Figure~\ref{fig: model} (c), our generator is diffusion-based, incorporating a denoising objective:
\begin{equation}
\mathcal{L} _{simple}=\mathbb{E} _{X^0,t,\epsilon}\left[ \lambda _t\left\| X-D_{\theta}\left( X+\sigma _t\epsilon , t, c \right) \right\| _{2}^{2} \right] 
\end{equation}
where $X^0$  denotes the ground-truth motion representation. Beyond the core denoising loss $\mathcal{L} _{simple}$, additional losses are used to ensure smooth and natural motion generation. Following prior work~\cite{tevet2023mdm, liang2024intergen}, we include joint velocity loss $\mathcal{L}_{vel}$ to enforce temporal consistency, foot joint contact loss $\mathcal{L}_{foot}$ to model foot-ground interactions, bone length loss $\mathcal{L}_{BL}$ to preserve skeletal proportions, and relative orientation loss $\mathcal{L}_{RO}$ to maintain proper facing direction during two-person interactions. For detailed formulations of these losses, see~\cite{tevet2023mdm, liang2024intergen} and our code repository.

\textbf{MC \& MI Losses} Human motion generative models are typically trained to produce composite representations that combine joint positions, velocities, and rotations. To facilitate better convergence during training, we design and adopt the following losses. While the simple loss $\mathcal{L} _{simple}$ enables the model to capture the overall motion representation, it neglects internal consistency. As demonstrated in Section~\ref{sec: experiments}, the three components (position, velocity, rotation) of the model’s \textbf{generated motion representations} diverge significantly. Specifically, within a motion representation $X$, the position and rotation components represent distinct motions, and the velocity component further deviates from the other two components. To address this, we introduce the \textbf{M}otion \textbf{C}onsistency loss, defined as
\begin{equation}
    \mathcal{L} _{MC}=\left\| \tilde{v}_{1:T-1}-\left( x_{2:T}-x_{1:T-1} \right) \right\| _{2}^{2}+\left\| \tilde{x}_{1:T}-FK\left( r_{1:T} \right) \right\| _{2}^{2}
\end{equation}
where $x$, $v$ and $r$ denote joint positions, velocities and rotations, respectively. The $\tilde{\cdot}$ means ground truth. And the $FK(\cdot)$ computes forward kinematics, mapping joint rotations to positions. 

\begin{figure}[h]
    \centering
    \includegraphics[width=0.99\linewidth]{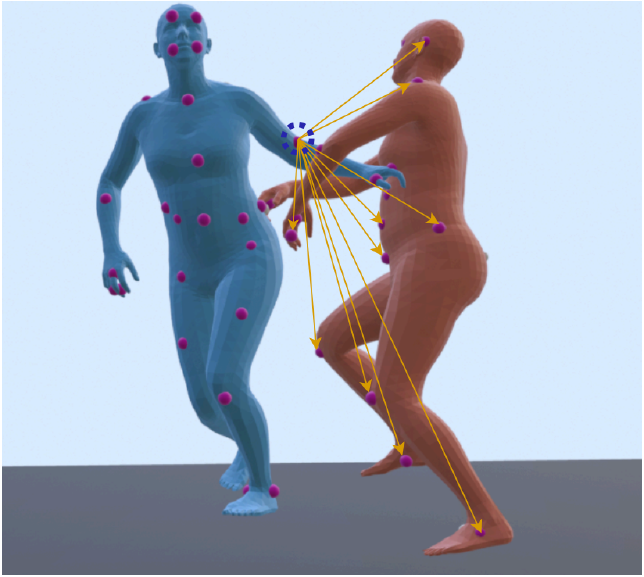}
    \caption{\textbf{Marker-Based Interaction Loss} The pink points represent pre-defined motion capture (mocap) markers, with a total of 67 markers per human. The proposed marker interaction (MI) loss computes pairwise distances between markers to construct a $67\times67$ distance matrix. This matrix is used to learn human interaction patterns during motion generation.}
    \label{fig: marker}
\end{figure}

Human interaction is mediated by the body’s surface. Prior work often relied on joint-point distances to model interactions, which oversimplifies real-world scenarios. Using mesh vertices for loss computation, however, incurs excessive computational overhead and redundant information, hindering model learning. To balance fidelity and efficiency, we propose the \textbf{M}arker-based \textbf{I}nteraction (MI) loss:
\begin{equation}
    \begin{split}
        \mathcal{L} _{MI}&=\left\| \mathbb{M} \left( m_{1:T}^{a}, m_{1:T}^{b} \right) \odot \left( \mathbb{M} \left( \tilde{m}_{1:T}^{a}, \tilde{m}_{1:T}^{b} \right) <0.1 \right) \right\| _{2}^{2}\\
        &+\| \left( \mathbb{M} \left( m_{1:T}^{a}, m_{1:T}^{b} \right) -\mathbb{M} \left( \tilde{m}_{1:T}^{a}, \tilde{m}_{1:T}^{b} \right) \right) \\
        &\odot \left( \mathbb{M} \left( m_{1:T}^{a}, m_{1:T}^{b} \right) <1 \right) \| _{2}^{2}
    \end{split}
    \label{eq: miloss}
\end{equation}
As illustrated in Figure~\ref{fig: marker}, markers $m$ are sampled from human meshes, derived from mocap markers. As discussed in prior work~\cite{wang2023refit, yao2023whmr}, mocap markers are sparse yet geometrically representative points, outperforming random or semantic sampling. $\mathbb{M}$ denotes a $67\times67$ distance map distance map computed from pairwise marker distances between humans $a$ and $b$. The operator $\odot$ applies element-wise masking, which is used to mask out the parts that do not meet the conditions. As shown in Eq.~\ref{eq: miloss}, the MI loss comprises two terms. The first term enforces contact when marker distances fall below 0.1 m. The second term aligns predicted distances with ground truth when distances are below 1 m, as interactions beyond this threshold are negligible.

\section{Experiments}\label{sec: experiments}
\subsection{Implementation Details}
Our method imposes no inherent limitations on application scope. It is agnostic to the choice of motion imitation technique and is compatible with diverse motion capture datasets and generative models. To evaluate its efficacy, we integrate our approach with the latest physics-based motion imitation method, PHC~\cite{luo2023phc}, using its position-based variant. To comprehensively assess the impact of physical mapping on motion generation, we conduct experiments on two benchmark datasets~\cite{liang2024intergen, guo2022generating} and compare against two classic baselines~\cite{liang2024intergen, tevet2023mdm}.
We prioritize multi-person interaction generation as the primary task, as it inherently involves both human-human and human-ground interactions. So, it is more suitable for validating our method. In addition to physical mapping, our proposed innovations (a novel human interaction representation, MC loss, and MI loss) are incorporated into this task. To further demonstrate the generalizability of physical mapping, we extend experiments to single-person text-to-motion generation, which is a foundational task in motion synthesis.
All experiments are conducted on an NVIDIA RTX 3090/4090 GPU, deployed under Linux using PyTorch. Physical simulations are executed within NVIDIA Isaac Gym~\cite{makoviychuk2021isaac}, a GPU-accelerated platform optimized for high-fidelity physics.

\subsubsection{\textbf{Datasets}}Our experiments focus on two tasks: multi-person interaction generation and single-person motion generation. We compare against two baselines InterGen~\cite{liang2024intergen} and MDM~\cite{tevet2023mdm}, which utilize the InterHuman~\cite{liang2024intergen} and HumanML3D~\cite{guo2022generating} datasets, respectively. Accordingly, our experiments are conducted on these datasets.

\textbf{InterHuman} The InterHuman dataset~\cite{liang2024intergen} captured two-person interactions using 76 calibrated Z-CAM RGB cameras, yielding 107 million frames across 7,779 motion sequences (6.56 hours total). An off-the-shelf pipeline~\cite{li2021ai} was employed to extract skeletal motion from RGB data. Text annotations were crowdsourced via Amazon Mechanical Turk, segmenting interactions into discrete 10-second clips to preserve semantic coherence. The dataset contains 23,337 unique descriptions (5,656 distinct words) spanning everyday interactions (e.g., hugs, handshakes) and specialized motions (e.g., dance, martial arts).

\textbf{HumanML3D} HumanML3D~\cite{guo2022generating} is a large-scale dataset for 3D human motion synthesis, comprising 14,616 motion sequences and 44,970 text descriptions. Each sequence is annotated with three distinct descriptions crowdsourced via Amazon Mechanical Turk to ensure diversity. Motions are represented in a canonical format. For our experiments, joint positions are used for physical mapping and subsequently converted to this format for model training.

\subsubsection{\textbf{Metrics}} Evaluating the quality of generated motion remains an unresolved challenge. Prior work has adapted metrics from text-to-image generation, which rely on feature extraction. However, current motion feature extraction methods lack sufficient data and robust models, often failing to capture motion quality accurately. These limitations have been noted in prior studies~\cite{tseng2023edge, meng2024rethinking}. A critical oversight is the direct evaluation of raw motion representations from models, which conflates inconsistent position, velocity, and rotation components, leading to unreliable results. Additionally, the physical plausibility of human motion lacks systematic assessment. To address these gaps, we propose the following metrics for comprehensive evaluation:

\begin{figure}[h]
    \centering
    \includegraphics[width=0.92\linewidth]{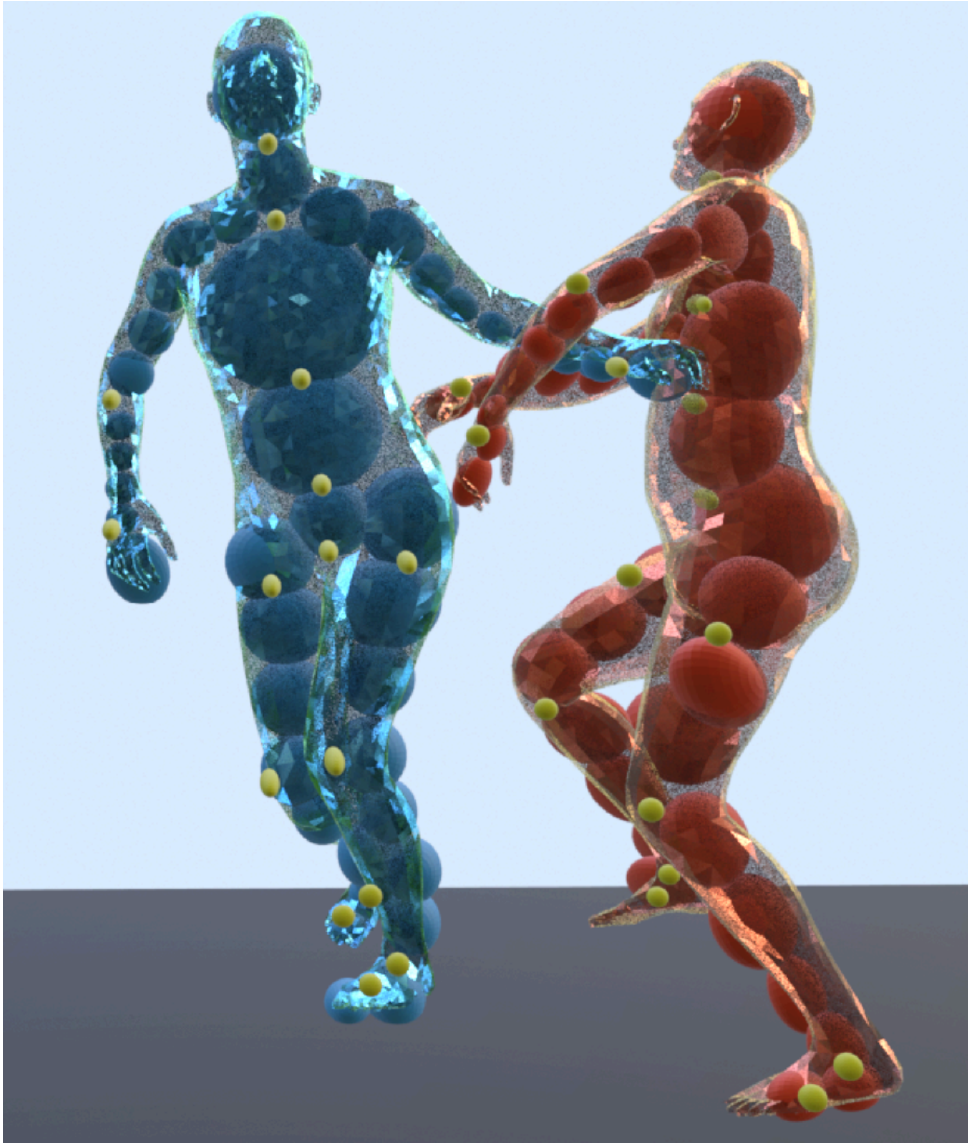}
    \caption{\textbf{Visualization of Human Mesh Simplification} To reduce computational complexity for interpenetration volume estimation, the human body mesh is approximated as a geometric structure composed of 45 spheres. And the yellow points here are joints.  }
    \label{fig: Interpenetration}
\end{figure}

\begin{table*}[h]
\renewcommand{\arraystretch}{1.5}
\caption{Comparison with baselines in terms of physical plausibility. Datasets include InterHuman~\cite{liang2024intergen} and HumanML3D~\cite{guo2022generating}; generative models include InterGen~\cite{liang2024intergen} and MDM~\cite{tevet2023mdm}. Ours denotes baselines integrated with our method, while boldface indicates superior performance of our approach.}
\setlength{\tabcolsep}{1.8mm}{
\begin{tabular}{cccccccc}
\hline
Task                                            & Dataset \& Method & Penetration $\downarrow \, (mm)$ & Float $\downarrow \, (mm)$ & Foot Contact $\downarrow \, (mm)$ & Skate $\downarrow \, (cm/s)$ & PFC $\downarrow$   & Interpenetration $\downarrow \, (cm^3)$ \\ \hline
\multirow{5}{*}{\rotatebox{90}{multi-person}}  & InterHuman     & 52.6795          & 8.6997    & 61.3792          & 24.5190     & 8.4764  & 20.2815                                 \\
                                                &Ours    & \textbf{18.9108}          &9.9969    & \textbf{28.9076}          & \textbf{13.4783}     & \textbf{3.8844}  & \textbf{14.8296}                                 \\
                                                &\cellcolor[HTML]{EFEFEF} InterGen-Pos   & \cellcolor[HTML]{EFEFEF}73.4320          & \cellcolor[HTML]{EFEFEF}1.3810    & \cellcolor[HTML]{EFEFEF}74.8138          &\cellcolor[HTML]{EFEFEF} 22.9522     &\cellcolor[HTML]{EFEFEF} 13.7305 &\cellcolor[HTML]{EFEFEF} 170.3378                                \\
                                                &\cellcolor[HTML]{EFEFEF} InterGen-Rot   &\cellcolor[HTML]{EFEFEF} 71.6635          &\cellcolor[HTML]{EFEFEF} 1.9317    &\cellcolor[HTML]{EFEFEF} 73.5952          &\cellcolor[HTML]{EFEFEF} 39.6250     &\cellcolor[HTML]{EFEFEF} 20.6950 &\cellcolor[HTML]{EFEFEF} 227.4542                                \\
                                                &\cellcolor[HTML]{EFEFEF} Ours      &\cellcolor[HTML]{EFEFEF} \textbf{6.6956}           &\cellcolor[HTML]{EFEFEF} 5.6492   &\cellcolor[HTML]{EFEFEF} \textbf{12.3447}          &\cellcolor[HTML]{EFEFEF} \textbf{9.3477}     &\cellcolor[HTML]{EFEFEF} \textbf{6.2246}  &\cellcolor[HTML]{EFEFEF} \textbf{95.5648} \\ \hline
\multirow{4}{*}{\rotatebox{90}{single-person}} & HumanML3D      & 3.1509           & 33.0275   & 36.1784          & 1.2962      & 6.4327  & -                                       \\
                                                & Ours     & 4.5257           & \textbf{17.1482}   & \textbf{21.6740}          & 3.0985      & \textbf{2.0291}  & -                                       \\
                                                &\cellcolor[HTML]{EFEFEF} MDM            &\cellcolor[HTML]{EFEFEF}1.3565                  &\cellcolor[HTML]{EFEFEF}62.7265          &\cellcolor[HTML]{EFEFEF}64.0930                  & \cellcolor[HTML]{EFEFEF}0.5622            &\cellcolor[HTML]{EFEFEF}10.5469         &\cellcolor[HTML]{EFEFEF} -                                       \\
                                                &\cellcolor[HTML]{EFEFEF} Ours           &\cellcolor[HTML]{EFEFEF}\textbf{3.4764}                  &\cellcolor[HTML]{EFEFEF}\textbf{5.3261}           &\cellcolor[HTML]{EFEFEF}\textbf{8.8024}                  &\cellcolor[HTML]{EFEFEF}\textbf{4.6395}             &\cellcolor[HTML]{EFEFEF}\textbf{2.2811}         &\cellcolor[HTML]{EFEFEF} -                                       \\ \hline
\end{tabular}}
\label{tab: phys}
\end{table*}

\textbf{Physical Metrics} A core contribution of this work is enhancing the physical plausibility of generated motions, a dimension not captured by feature-based metrics. To quantify physical plausibility, we introduce the following metrics: (1) \textit{Penetration}: The average depth by which the human mesh penetrates the ground plane. (2) \textit{Float}: The average height of the lowest point of the human mesh above the ground. (3) \textit{Foot Contact}: Defined as $Penetration + Float$, this metric evaluates the interaction between the human mesh and the ground. (4) \textit{Skate}: The average horizontal sliding velocity of the human model. This metric only concerns the part of the body that penetrates the ground. (5) \textit{Physical Foot Contact Score (PFC)}:  A metric inspired by prior work~\cite{tseng2023edge, zheng2024beat} to assess sliding. PFC evaluates whether foot move coincides with body acceleration, as simultaneous sliding and acceleration disrupt balance. PFC is calculated as  $PFC=v_{foot}^{left}\cdot v_{foot}^{right}\cdot a_{root}$, where $v_{foot}$ denotes foot velocity and $a_{root}$ is the root joint acceleration. Smaller PFC values indicate reduced foot sliding during human motion. (6) \textit{Interpenetration}: The average volumetric overlap between human meshes in multi-person scenarios. As shown in Figure.~\ref{fig: Interpenetration}, we approximate the human body as 45 spheres and compute the intersecting volume between spheres as a coarse interpenetration measure. (7) \textit{Mean Per-Joint Position Error (MPJPE)}: A novel metric comparing joint positions between position-based and rotation-based motion representations. This metric highlights discrepancies between motion components, underscoring the need for decoupled evaluation and application.

\textbf{Conventional Metrics} Following prior work \cite{guo2022generating, guo2020action2motion, lee2019dancing}, we acknowledge the limitations of feature-based metrics yet retain them for comparative evaluation, reporting results for reference. Our assessment distinguishes between position-based and rotation-based metrics. For position-based evaluation, velocity and rotation are derived from the positional components of motion data, forming a synthesized motion representation for analysis. But for rotation-based evaluation, positions are derived from rotation and then velocity is calculated from positions. The specific metrics include: (1) \textit{R-Precision}: Quantifies the alignment between generated motion sequences and text prompts. (2) \textit{FID}: Measures divergence between the distribution of generated motions and real motion data. (3) \textit{MM Dist}: Evaluates the cross-modal dissimilarity between motion and text representations. (4) \textit{Diversity}: Assesses the variability of generated motions via statistical feature distribution analysis. (5) \textit{MModality}: Unlike Diversity, which captures all motion distribution, MModality evaluates the diversity of motions conditioned on the same text prompts. (6) \textit{FID*}: The difference between FID* and FID is that FID* directly uses $22\times3$ joint positions to calculate, while FID requires additional feature extractors to extract features.

\subsection{Evaluation Results} 

\subsubsection{\textbf{Physical Plausibility}} A central contribution of this work is improving the physical plausibility of generated motions. As detailed in Table~\ref{tab: phys}, our experiments evaluate both single-person and multi-person scenarios. For multi-person motion generation, we utilize the InterHuman dataset. The first three metrics in the table reveal significant ground penetration in the original dataset. In contrast, the augmented InterHuman dataset exhibits a 64\% reduction in Penetration, with Float remaining relatively unchanged. This indicates enhanced foot-ground contact in the improved dataset. Additionally, our method significantly mitigates sliding artifacts. Specifically, the Skate metric (overall sliding) and PFC (a refined sliding measure) demonstrate 45\% and 54\% reductions, indicating motions that better adhere to real-world physics. For multi-person scenarios, interpersonal penetration is critical. Under identical motion samples, the enhanced dataset shows a 27\% reduction in Interpenetration, further validating the physical realism of our approach.

The same phenomenon is also evident in single-person experiments. We evaluate our method using the HumanML3D dataset, the most widely adopted text-to-motion benchmark. And we compare our method against the classic MDM model. Our approach significantly outperforms the baseline on most metrics, with the exception of Penetration and Skate. This outcome is attributed to the original dataset’s motions being generally positioned above the ground plane, leading to minimal penetration and sliding velocity. However, results from the comprehensive metrics Foot Contact and PFC demonstrate that our physical mapping indeed enhances human-ground interaction, rendering generated motions more physically plausible.

\begin{figure}[ht]
    \centering
    \includegraphics[width=\linewidth]{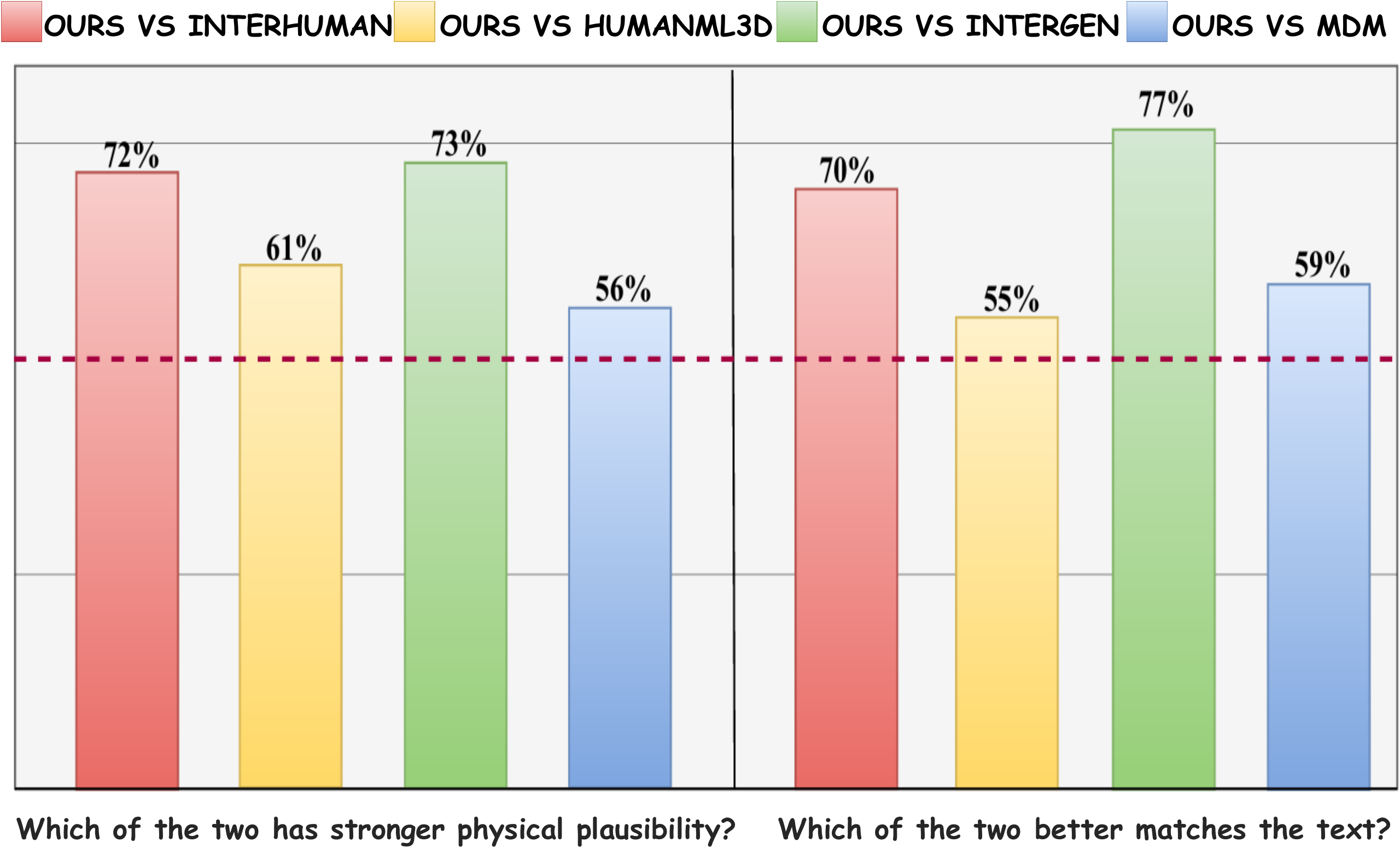}
    \caption{\textbf{User Study Results} We use a scoring system to compare ours with various benchmark datasets and methods, using physical plausibility and text matching as the criteria. The red dashed line indicates 50\%. Exceeding the red line means ours is preferred.}
    \label{fig: user study}
\end{figure}

\begin{figure}[ht]
    \centering
    \includegraphics[width=\linewidth]{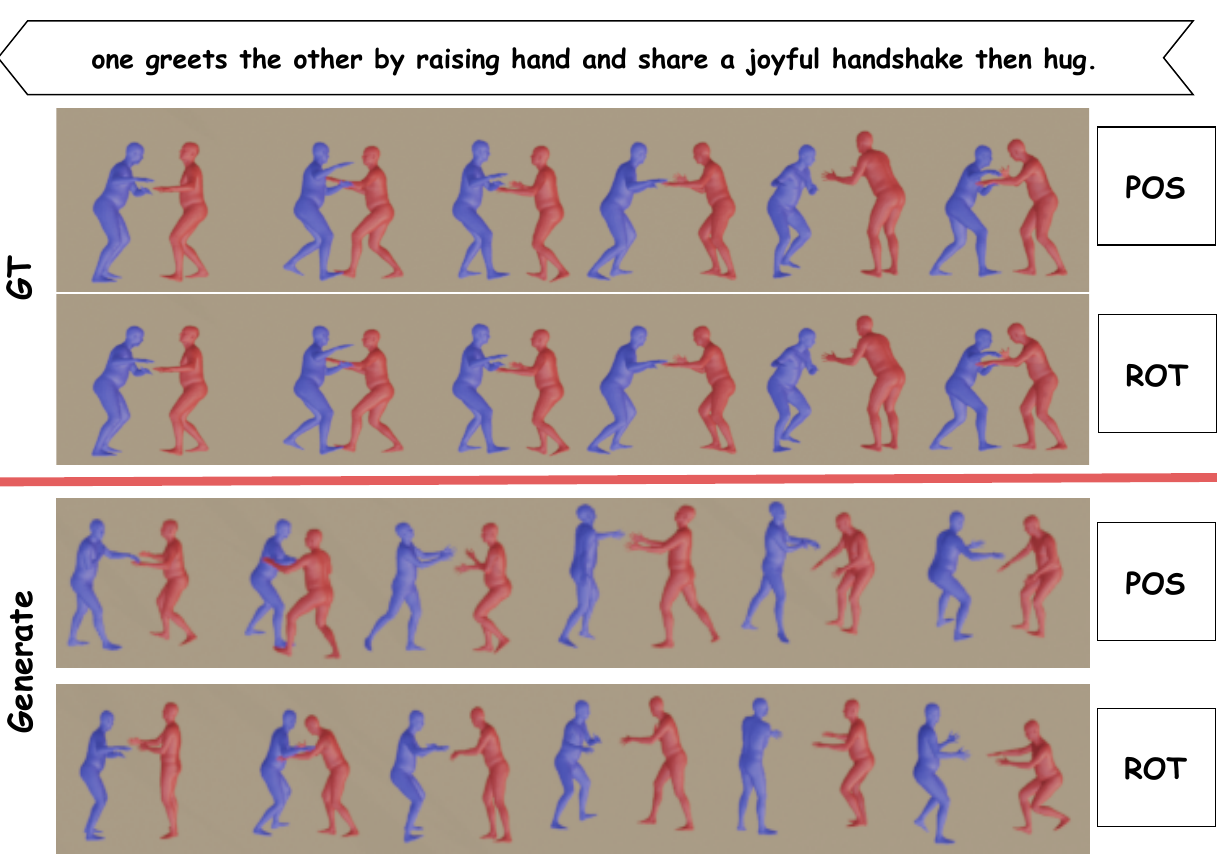}
    \caption{\textbf{Comparison of Position-based and Rotation-based} \textit{GT} represents the ground-truth motion representation in the dataset. \textit{Generate} represents the motion generated by the model. \textit{POS} and \textit{ROT} represent position-based and rotation-based motions.}
    \label{fig: pos_and_rot}
\end{figure}

\subsubsection{\textbf{User Study}}
To evaluate the perceptual quality of generated motions, we conducted a controlled user study comparing our method against four datasets and methods: InterHuman, HumanML3D, InterGen, and MDM. The study involved 50 participants. And each participant evaluated 10 randomized motion pairs (ours vs. baseline) under two criteria. First, \textbf{Physical Plausibility} assesses biomechanical feasibility and interpenetration. Second, \textbf{Text-Motion Alignment} judges semantic consistency between motion sequences and text descriptions. As demonstrated in Figure~\ref{fig: user study}, our method achieves superior preference rates across all comparison groups. The performance gap widens significantly in complex interaction scenarios. 70+\% of participants preferred our results in multi-person motions, while only 55+\% of participants chose our results when it comes to single-person motion.

\begin{figure*}[hbp]
    \centering
    \includegraphics[width=0.9\textwidth]{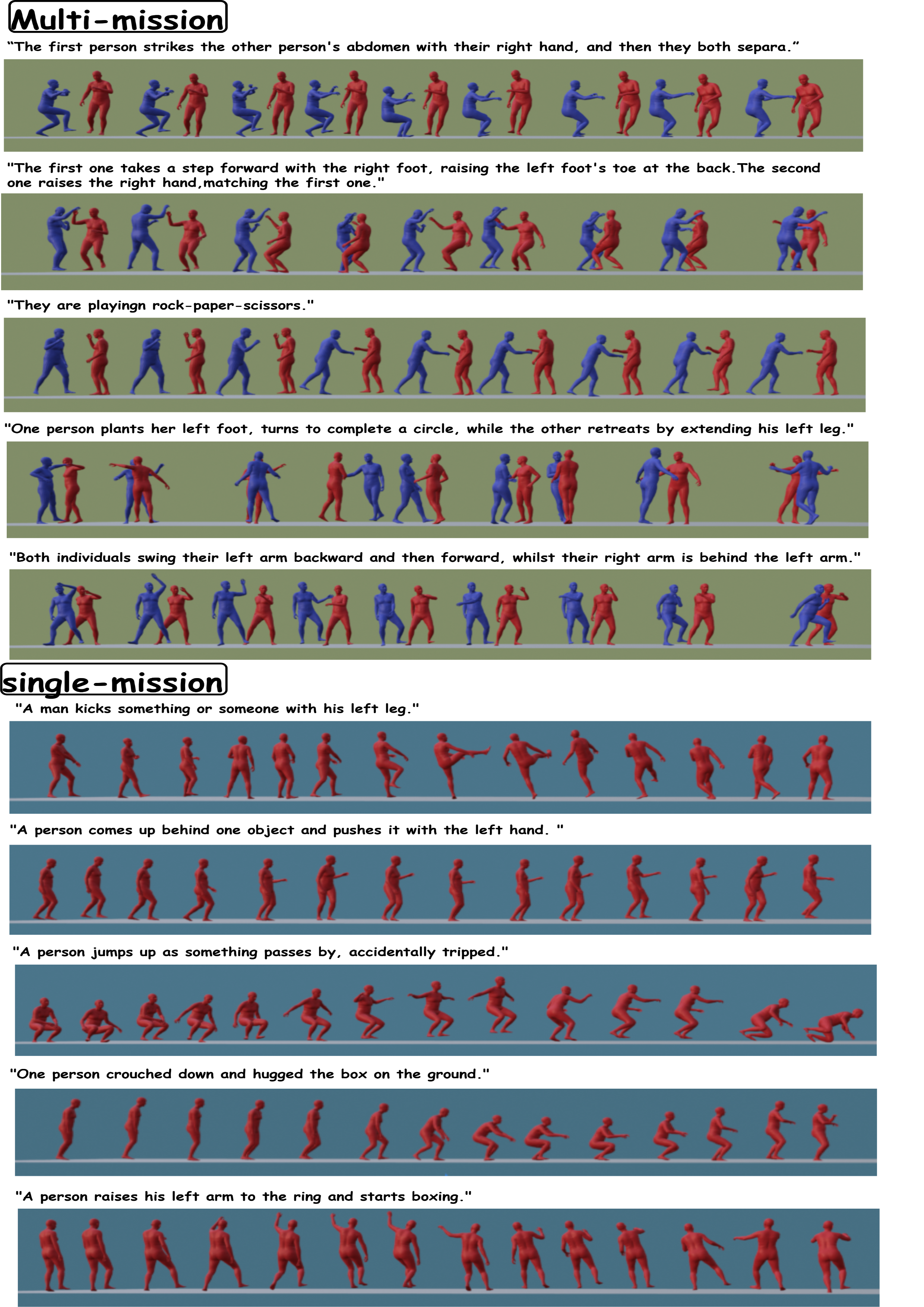}
    \caption{\textbf{Qualitative Results} We present qualitative results for both single-person and multi-person motion generation. Our method produces motion sequences that align with text prompts while maintaining high physical plausibility.}
    \label{fig: bigpic}
\end{figure*}

Three key insights emerge from the study. Physical plausibility serves as a perceptual prerequisite for text-motion alignment evaluation. Multi-person interactions amplify the benefits of physics-based generation. Local contact refinement significantly impacts global motion perception. Surprisingly, our method also surpasses all others in text-motion matching. The results validate our core hypothesis: explicit physical modeling not only improves motion quality but also enhances perceived semantic alignment through biomechanically plausible interactions. Please refer to Figure~\ref{fig: bigpic} and supplementary materials for more qualitative results.

\begin{table}[]
\caption{Comparison of different motion representations. InterGen means that the model output results are directly used during evaluation. InterGen-Pos means that the velocity part and rotation part are corrected according to the position part of the output results before evaluation. InterGen-Rot calculates the position part and velocity part based on the rotation part.}
\label{tab:rep_comp}
\begin{tabular}{ccccc}
\hline
Methods      & FID $\downarrow$ & MM Dist $\downarrow$ & Diversity $\uparrow$ & MModality $\uparrow$ \\ \hline
InterGen     & 5.918            & 5.108                & 7.387                & 2.141                \\
InterGen-Pos & 33.229           & 3.877                & 7.318                & 1.092                \\
InterGen-Rot & 80.424           & 3.973                & 7.947                & 1.230                \\ \hline
\end{tabular}
\end{table}

\subsubsection{\textbf{Conventional Evaluation}} Conventional evaluation metrics for motion generation exhibit fundamental limitations: they predominantly assess motion distributions in a coarse-grained manner through low-dimensional statistical features, failing to capture fine-grained motion quality while heavily relying on the representational capacity of evaluators. Furthermore, suboptimal motion representations exacerbate measurement biases, as demonstrated in our first benchmark study. To address these limitations, we design three systematic experiments: (1)~a diagnostic analysis revealing significant inconsistencies among conventional metrics when evaluating identical motion sequences, (2)~a ablation study through our novel metric FID* demonstrating motion quality improvement, and (3)~a cross-modal alignment evaluation where our method achieves better motion-text semantic matching.

\textit{Representation Ambiguity in Motion Metrics.} The inherent instability of conventional metrics arises from their failure to address the fundamental discrepancy between position and rotation in the motion representation. As demonstrated in Tab.~\ref{tab:rep_comp}, identical motions expressed through different motion representation parts yield drastically divergent FID scores (5.918 vs. 33.229 vs. 80.424). The root cause of this phenomenon is that the generative model does not strictly constrain the internal consistency of the motion representation, resulting in different parts of the same motion representation representing different motions, as shown in Figure. ?. This representation ambiguity propagates through evaluation pipelines, as metrics like FID mechanically process contradictory geometric priors through identical feature extractors, conflating kinematic validity with parametric artifacts. A critical flaw exacerbated by InterGen-Pos/InterGen-Rot variants where reprojected motion representation amplifies inconsistencies, ultimately rendering scores non-comparable across studies using different motion parameterizations.  

\begin{figure}[ht]
    \centering
    \includegraphics[width=\linewidth]{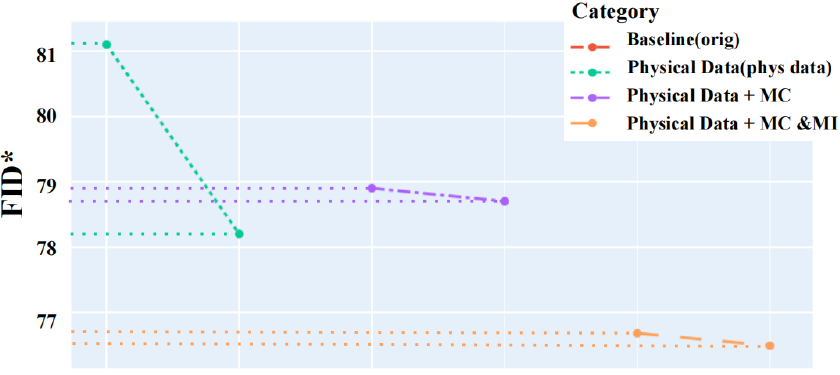}
    \caption{\textbf{Quantitative Results of Motion Quality.} FID* evaluates motion quality by comparing it with the motion distribution of real motion. The smaller the value, the better the motion quality. Each group has two values measured using position-based and rotation-based motion. The evaluation results of Baseline are 125.39091
    and 123.3034 respectively. They are not shown in the figure because of their large values.}
    \label{fig: fid}
\end{figure}

\textit{Improved FID* Metric Analysis.} As shown in Figure~\ref{fig: fid}, our proposed FID* metric circumvents feature extractor biases by directly computing distribution distances on 22$\times$3 joint positions, revealing critical insights: (1) InterGen variants show order-of-magnitude improvements from raw outputs through phased integration of physics-based data and kinematic consistency losses, with position/rotation discrepancy decreasing from $\Delta$2.09 to $\Delta$0.22 under our MC (Motion Consistency) loss, which is also validated by MPJPE in Table~\ref{tab: ablation}. This progression confirms that disentangling representation artifacts effectively elevates motion quality measurement objectivity. It is worth mentioning that we also tested the FID* of the dataset. The Real (augmented with physical mapping) is 0.0110, and the Real-ori (original dataset) is 0.5830. This shows that those physically impossible artifacts are not obvious from the perspective of data distribution. However, the generator will amplify such artifacts and cause a great impact. This once again emphasizes the necessity of using physical mapping to correct training data.

\begin{table*}
\renewcommand{\arraystretch}{1.25}
\centering
\caption{Ablation studies on InterHuman, evaluated in terms of physical plausibility.}
\label{tab: ablation}
\setlength{\tabcolsep}{3mm}{
\begin{tabular}{ccccccccc} 
\hline
\multicolumn{2}{c}{Methods}                                                                                                                                                                      & Penetration $\downarrow $ & Float $\downarrow $ & Foot Contact $\downarrow $ & Skate $\downarrow $ & PFC $\downarrow$   & Interpenetration $\downarrow $ & MPJPE $\downarrow$                                                          \\ 
\hline
\multicolumn{2}{c}{Real}                                                                                                                                                                          & 18.9108     & 9.9969  & 28.9076      & 13.4783 & 3.8844  & 14.8296          & 0                                                               \\ 
\hline
\rowcolor[rgb]{0.937,0.937,0.937} {\cellcolor[rgb]{0.937,0.937,0.937}}                                                                                                             & InterGen-Pos & 73.4320     & 1.3810  & 74.8138      & 22.9522 & 13.7305 & 170.3378         & {\cellcolor[rgb]{0.937,0.937,0.937}}                            \\
\rowcolor[rgb]{0.937,0.937,0.937} \multirow{-2}{*}{{\cellcolor[rgb]{0.937,0.937,0.937}}Baseline}                                                                                       & InterGen-Rot & 71.6635     & 1.9317  & 73.5952      & 39.6250 & 20.6950 & 227.4542         & \multirow{-2}{*}{{\cellcolor[rgb]{0.937,0.937,0.937}}121.6017}  \\
\multirow{2}{*}{Phys Data}                                                                                                                                                         & InterGen-Pos & 9.3234      & 7.6762  & 16.9996      & 6.9230  & 11.4806 & 134.6332         & \multirow{2}{*}{63.2580}                                        \\
                                                                                                                                                                                   & InterGen-Rot & 9.2992      & 8.3056  & 17.6048      & 10.8824 & 14.3556 & 145.1245         &                                                                 \\
\rowcolor[rgb]{0.937,0.937,0.937} {\cellcolor[rgb]{0.937,0.937,0.937}}                                                                                                             & InterGen-Pos & 10.4730     & 5.0830  & 15.5560      & 9.1385  & 10.5066 & 223.8249         & {\cellcolor[rgb]{0.937,0.937,0.937}}                            \\
\rowcolor[rgb]{0.937,0.937,0.937} \multirow{-2}{*}{{\cellcolor[rgb]{0.937,0.937,0.937}}\begin{tabular}[c]{@{}>{\cellcolor[rgb]{0.937,0.937,0.937}}c@{}}Phys Data\\MC\end{tabular}} & InterGen-Rot & 7.3465      & 5.8198  & 13.1663      & 7.8615  & 10.9523 & 234.2524         & \multirow{-2}{*}{{\cellcolor[rgb]{0.937,0.937,0.937}}14.2965}   \\
\multirow{2}{*}{\begin{tabular}[c]{@{}c@{}}Phys Data\\MC \& MI\end{tabular}}                                                                                                          & InterGen-Pos & 10.2939     & 4.8826  & 15.1765      & 9.8936  & 9.7655  & 230.6373         & \multirow{2}{*}{14.4756}                                        \\
                                                                                                                                                                                   & InterGen-Rot & 6.9641      & 5.7212  & 12.6853      & 8.4606  & 10.4578 & 239.4154         &                                                                 \\
\rowcolor[rgb]{0.937,0.937,0.937} \begin{tabular}[c]{@{}>{\cellcolor[rgb]{0.937,0.937,0.937}}c@{}}Phys Data\\MC \& MI\\Phys Map\end{tabular}                                          & Ours    & \textbf{6.6956}      & 5.6492  & \textbf{12.3477}      & 9.3477  & \textbf{6.2246}  & \textbf{95.5648}          & 0                                                               \\ 
\hline
\multicolumn{2}{c}{Real}                                                                                                                                                                          & 4.5257      & 17.1482 & 21.6740      & 3.0985  & 2.0291  & -                & -                                                               \\ 
\hline
\rowcolor[rgb]{0.937,0.937,0.937} Baseline                                                                                                                                             & MDM          & 1.3565      & 62.7265 & 64.0930      & 0.5622  & 10.5469 & -                & -                                                               \\
Phys Data                                                                                                                                                                          & MDM          & 1.1125      & 47.3667 & 48.4921      & 8.6794  & 12.3280 & -                & -                                                               \\
\rowcolor[rgb]{0.937,0.937,0.937} \begin{tabular}[c]{@{}>{\cellcolor[rgb]{0.937,0.937,0.937}}c@{}}Phys Data\\Phys Map\end{tabular}                                                 & Ours         & 3.4764      & \textbf{5.3261}  & \textbf{8.8024}       & 4.6395  & \textbf{2.2811}  & -                & -                                                               \\
\hline
\end{tabular}}
\end{table*}

\begin{figure}[ht]
    \centering
    \includegraphics[width=\linewidth]{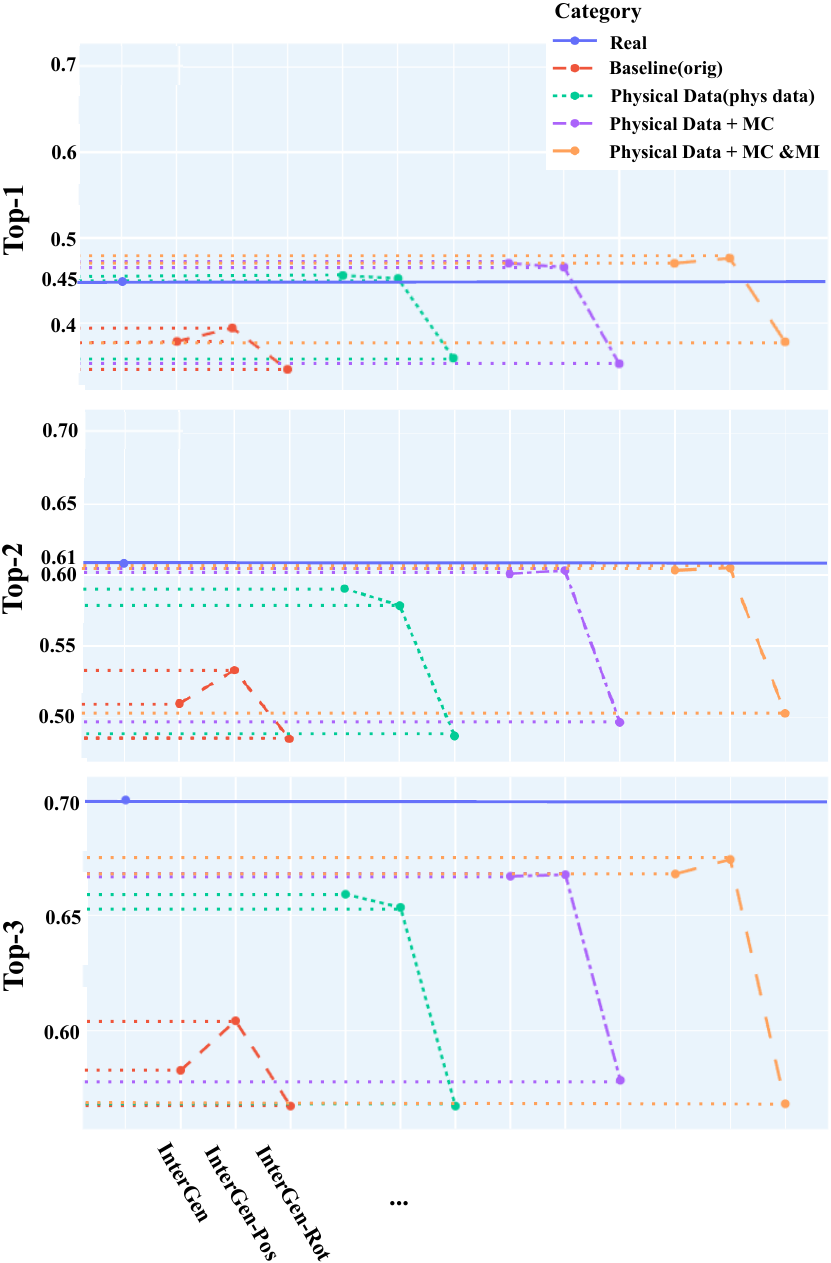}
    \caption{R-Precision performance on the InterHuman dataset. Experimental settings align with those described in Table~\ref{tab: ablation}.}
    \label{fig: r-precision}
\end{figure}

\textit{Text-Motion Alignment Validation.} As shown in Figure~\ref{fig: r-precision}, our retrained R-Precision evaluator on augmented InterHuman demonstrates systematic improvements across text-conditioned generation variants. This advancement stems from our dual strategy: (1)~physics-aware representation canonicalization through marker-based constraints, reducing pose-text ambiguity, and (2)~motion consistency (MC) regularization that minimizes KL-divergence between cross-representation feature distributions. Of course, the most fundamental improvement comes from the dataset enhanced with physical mapping, which lays an absolute foundation for training high-quality motion generators.

\subsection{Ablation Studies}\label{sec: ablation}
To validate the effectiveness of our proposed techniques, we conducted ablation studies using physical metrics. These experiments address three key questions: (1) Does training on datasets with enhanced physical plausibility improve the quality of generated motions? (2) Do the proposed MC and MI losses achieve their intended objectives? (3) What is the impact of post-processing via physical mapping on motion quality? The following section details our results and analysis.

\subsubsection{\textbf{Augmented Data}} As shown in Table~\ref{tab: ablation}, comparing the “Baseline" and “Phys Data" experiments reveals that augmented data significantly enhance motion generation quality. Specifically, metrics improve by 16\%–87\%, with the exception of \textit{Float}. The \textit{Float} metric remains stable because the baseline model tends to generate motions with excessive ground penetration (in multi-person scenarios) or excessive floating (in single-person scenarios). This behavior aligns with the dataset evaluation results in Table~\ref{tab: phys}, where the original InterHuman and HumanML3D datasets exhibit similar artifacts, i.e., ground penetration in multi-person interactions and floating in single-person motions. These findings underscore that dataset quality directly determines model performance: generative models often replicate dataset errors. Our results confirm that improving dataset quality via physical mapping is a direct and effective strategy to enhance motion generation performance.

\subsubsection{\textbf{MC \& MI Losses}} The MC loss is designed to address inconsistencies within the same motion representation. It serves two primary purposes: constraining the feature space to facilitate model training and broadening the applicability of generated motion representations, enabling their direct use for both joint positions and rotations. As shown in the “Baseline" and “Phys Data" groups of Table~\ref{tab: ablation}, we evaluate the MPJPE between position-based and rotation-based motions, which should ideally be zero. Generally, an MPJPE exceeding 50 $mm$ indicates a noticeable difference in motions. The evaluation results demonstrate that, without the MC loss, the rotation-based motions significantly diverge from the position-based ones, resulting in lower quality. The MC loss reduces the MPJPE to below 15 $mm$, making it nearly impossible to distinguish between the two motions. Consequently, the rotation-based motions achieve comparable quality to the position-based ones and can be directly utilized.

The MI loss is employed to train the model to learn more refined movements and human interactions. The marker-based loss enables the model to efficiently learn the subtle geometric changes of the human body without increasing training computational load. As can be seen from Tab.~\ref{tab: ablation}, $8/13$ metrics have been improved. Oddly, \textit{Interpenetration} slightly deteriorates. However, by analyzing Func.~\ref{eq: miloss}, it can be found that the first part encourages human-human contact. We believe that it is this factor that makes the interpenetration more severe. Nevertheless, we still consider the overall impact of the MI loss to be positive and retain it. 

\subsubsection{\textbf{Post Processing}} This section focuses on the role of physical mapping in post-processing. Its contribution to data augmentation is discussed earlier. As illustrated in Table~\ref{tab: ablation}, physical mapping is applied to two motion generation tasks, yielding significant improvements. Physical mapping serves as an explicit constraint, ensuring motions adhere to physical laws. Both mocap data and generated motions conform to physical constraints after physical mapping. Its consistent effectiveness across diverse tasks and models demonstrates our method’s generalizability. Performance gains of 3\%–89\% across multiple metrics further validate its utility.

\section{Discussion}

\subsection{Limitation} 

Our proposed method is fundamentally based on motion imitation. Consequently, the inherent limitations of motion imitation techniques represent the primary constraints of our approach. Specifically, even SOTA motion imitation algorithms are unable to perfectly replicate all human motions. This limitation is particularly pronounced in scenarios involving complex human-human and human-scene interactions, where the likelihood of imitation failure increases significantly. The origins of these limitations can be attributed to three principal factors, which are detailed in the following sections.

\begin{figure}[ht]
    \centering
    \includegraphics[width=\linewidth]{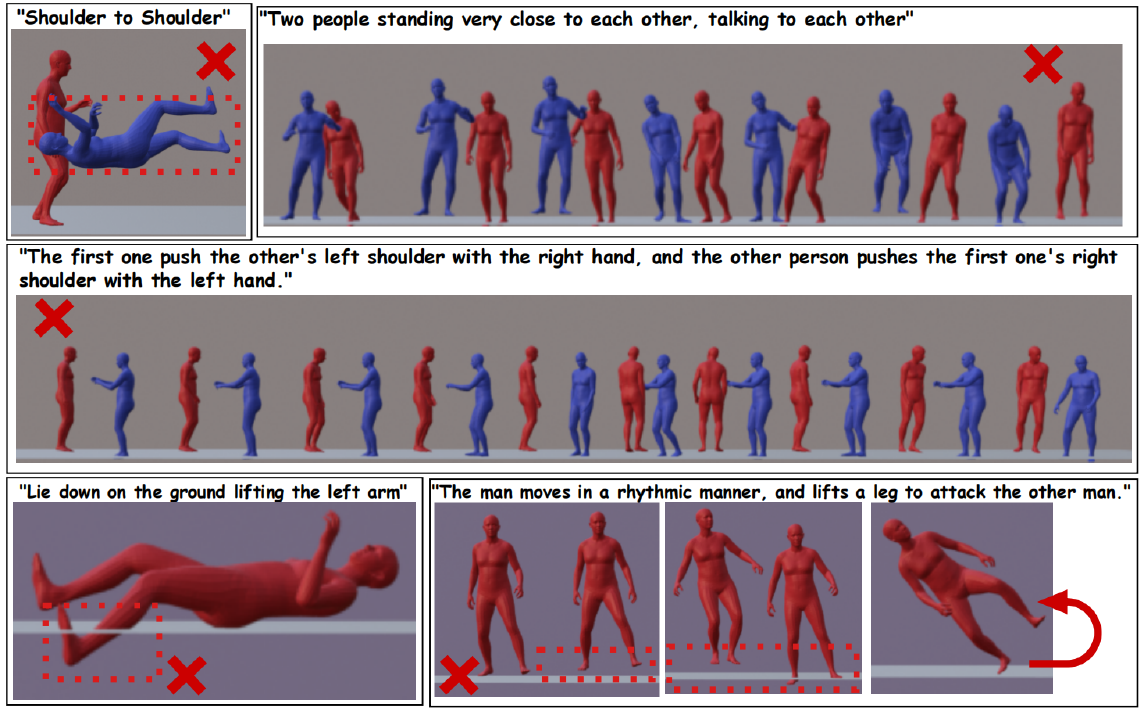}
    \caption{\textbf{Failure Cases} Limitations in our method lead to occasional failures, including falls, unnatural movements, incorrect collisions, and interpenetration artifacts.}
    \label{fig: failure}
\end{figure}

First, the imitation policy falls short in covering complex motions, such as backflips and handstands. As shown in Figure~\ref{fig: failure}, a significant number of the failed cases involve agents falling, which can be attributed to this limitation. This highlights two main issues. On one hand, there is a need for further improvement in the policy design. On the other hand, the scarcity of relevant training data hinders the model's ability to be trained effectively for such challenging scenarios. 

Second, the human model employed by the agent does not accurately represent the real human body. Real humans possess a highly diverse geometric shape, while in the simulator, it is oversimplified as a combination of several spheres and cubes. This simplification clearly fails to simulate the contact between humans and the external environment. As shown in Figure~\ref{fig: failure}, the unrealistic penetration and gaps during contact are a direct result of the discrepancies between the shapes of the human body and the simulated agent. Third, the simulator's representation of the physical world is imperfect. Factors such as gravity, friction, collision detection, and the physical properties of humans all have an impact on the simulation results. This imperfection can lead to distortion of many imitated motions. For example, the falling speed of a person and the reaction during a collision may appear unnatural. More seriously, it can make some interactive motions extremely hard to imitate. An example is the motion of one person picking up another person, which involves complex physical interactions.

\subsection{Future Work} 

Based on the analysis presented above, our future research should focus on further exploring the potential of motion imitation. Targeted improvements can be made to address the three limitations identified earlier. First, a dataset of complex motions can be constructed to train an imitation policy capable of covering a broader range of motions. Second, a more detailed human body model can be developed, aligning more closely with real human geometry, to enable agents to simulate human interactions with greater accuracy. Third, the physical properties of the simulator can be refined to more closely approximate real-world conditions, thereby providing a foundation for imitating complex interactive motions. As a critical bridge between the virtual and physical domains, motion imitation plays a crucial role in advancing spatial intelligence and embodied intelligence. This area warrants increased research efforts and investment.

In addition to motion imitation, numerous unresolved challenges in motion generation demand attention. This paper introduces a motion representation better suited to task requirements and proposes more effective loss functions. Furthermore, we developed a more objective and robust evaluation framework. Despite these contributions, several key questions remain open: What constitutes the optimal motion representation for generative models? What loss function most effectively constrains the generation of realistic human motions? How can the realism and text alignment of generated motions be objectively assessed? Addressing these questions is critical to advancing motion generation. We hope this work will stimulate further research and inspire innovative solutions within the community.

\section{Conclusion}

In this paper, we presented a series of methods to enhance the physical plausibility of human interaction generation. We introduced physical mapping, a technique that enforces physical constraints on human motions through motion imitation within a physics simulator.
First, we employed physical mapping to augment the motion capture dataset. This process enriched the motion distribution learned by the generative model, enabling it to produce more realistic human motions. Second, we applied physical mapping as a post-processing step for generated motions. By directly enforcing physical constraints, this method ensured the physical validity of the output.
Additionally, we proposed a motion representation tailored for generative models, which facilitates both learning and application. The introduced MC loss enhances training efficiency and ensures that all components of the generated results meet task-specific criteria. The MI loss further improves the model’s ability to generate human-human interactions.
Finally, we developed a systematic evaluation framework. Extensive experiments demonstrate that our methods significantly enhance the physical plausibility of generated motions. We also validated the approach on various human motion generation tasks, confirming its generalizability. We hope this work will inspire further advancements in the community.

\bibliographystyle{IEEEtran}

\bibliography{main}

\begin{thebibliography}{10}
\providecommand{\url}[1]{#1}
\csname url@samestyle\endcsname
\providecommand{\newblock}{\relax}
\providecommand{\bibinfo}[2]{#2}
\providecommand{\BIBentrySTDinterwordspacing}{\spaceskip=0pt\relax}
\providecommand{\BIBentryALTinterwordstretchfactor}{4}
\providecommand{\BIBentryALTinterwordspacing}{\spaceskip=\fontdimen2\font plus
\BIBentryALTinterwordstretchfactor\fontdimen3\font minus \fontdimen4\font\relax}
\providecommand{\BIBforeignlanguage}[2]{{%
\expandafter\ifx\csname l@#1\endcsname\relax
\typeout{** WARNING: IEEEtran.bst: No hyphenation pattern has been}%
\typeout{** loaded for the language `#1'. Using the pattern for}%
\typeout{** the default language instead.}%
\else
\language=\csname l@#1\endcsname
\fi
#2}}
\providecommand{\BIBdecl}{\relax}
\BIBdecl

\bibitem{pymaf}
H.~Zhang, Y.~Tian, X.~Zhou, W.~Ouyang, Y.~Liu, L.~Wang, and Z.~Sun, ``Pymaf: 3d human pose and shape regression with pyramidal mesh alignment feedback loop,'' in \emph{Proceedings of the IEEE/CVF International Conference on Computer Vision}, 2021, pp. 11\,446--11\,456.

\bibitem{pymaf-x}
H.~Zhang, Y.~Tian, Y.~Zhang, M.~Li, L.~An, Z.~Sun, and Y.~Liu, ``Pymaf-x: Towards well-aligned full-body model regression from monocular images,'' \emph{IEEE Transactions on Pattern Analysis and Machine Intelligence}, 2023.

\bibitem{yao2024staf}
W.~Yao, H.~Zhang, Y.~Sun, and J.~Tang, ``Staf: 3d human mesh recovery from video with spatio-temporal alignment fusion,'' \emph{IEEE Transactions on Circuits and Systems for Video Technology}, 2024.

\bibitem{vibe}
M.~Kocabas, N.~Athanasiou, and M.~J. Black, ``Vibe: Video inference for human body pose and shape estimation,'' in \emph{Proceedings of the IEEE/CVF conference on computer vision and pattern recognition}, 2020, pp. 5253--5263.

\bibitem{spec}
M.~Kocabas, C.-H.~P. Huang, J.~Tesch, L.~M{\"u}ller, O.~Hilliges, and M.~J. Black, ``Spec: Seeing people in the wild with an estimated camera,'' in \emph{Proceedings of the IEEE/CVF International Conference on Computer Vision}, 2021, pp. 11\,035--11\,045.

\bibitem{ho2020denoising}
J.~Ho, A.~Jain, and P.~Abbeel, ``Denoising diffusion probabilistic models,'' \emph{Advances in neural information processing systems}, vol.~33, pp. 6840--6851, 2020.

\bibitem{tevet2023mdm}
\BIBentryALTinterwordspacing
G.~Tevet, S.~Raab, B.~Gordon, Y.~Shafir, D.~Cohen-or, and A.~H. Bermano, ``Human motion diffusion model,'' in \emph{The Eleventh International Conference on Learning Representations}, 2023. [Online]. Available: \url{https://openreview.net/forum?id=SJ1kSyO2jwu}
\BIBentrySTDinterwordspacing

\bibitem{shafir2023priormdm}
Y.~Shafir, G.~Tevet, R.~Kapon, and A.~H. Bermano, ``Human motion diffusion as a generative prior,'' \emph{arXiv preprint arXiv:2303.01418}, 2023.

\bibitem{zhang2024motiondiffuse}
M.~Zhang, Z.~Cai, L.~Pan, F.~Hong, X.~Guo, L.~Yang, and Z.~Liu, ``Motiondiffuse: Text-driven human motion generation with diffusion model,'' \emph{IEEE Transactions on Pattern Analysis and Machine Intelligence}, 2024.

\bibitem{Zhang2023t2mgpt}
\BIBentryALTinterwordspacing
J.~Zhang, Y.~Zhang, X.~Cun, S.~Huang, Y.~Zhang, H.~Zhao, H.~Lu, and X.~Shen, ``Generating human motion from textual descriptions with discrete representations,'' \emph{2023 IEEE/CVF Conference on Computer Vision and Pattern Recognition (CVPR)}, pp. 14\,730--14\,740, 2023. [Online]. Available: \url{https://api.semanticscholar.org/CorpusID:255942203}
\BIBentrySTDinterwordspacing

\bibitem{Jiang2023MotionGPTHM}
\BIBentryALTinterwordspacing
B.~Jiang, X.~Chen, W.~Liu, J.~Yu, G.~Yu, and T.~Chen, ``Motiongpt: Human motion as a foreign language,'' \emph{ArXiv}, vol. abs/2306.14795, 2023. [Online]. Available: \url{https://api.semanticscholar.org/CorpusID:259262201}
\BIBentrySTDinterwordspacing

\bibitem{Tevet2022MotionCLIPEH}
\BIBentryALTinterwordspacing
G.~Tevet, B.~Gordon, A.~Hertz, A.~H. Bermano, and D.~Cohen-Or, ``Motionclip: Exposing human motion generation to clip space,'' in \emph{European Conference on Computer Vision}, 2022. [Online]. Available: \url{https://api.semanticscholar.org/CorpusID:247450907}
\BIBentrySTDinterwordspacing

\bibitem{Zhang2024motionmamba}
\BIBentryALTinterwordspacing
Z.~Zhang, A.~Liu, I.~Reid, R.~Hartley, B.~Zhuang, and H.~Tang, ``Motion mamba: Efficient and long sequence motion generation with hierarchical and bidirectional selective ssm,'' \emph{ArXiv}, vol. abs/2403.07487, 2024. [Online]. Available: \url{https://api.semanticscholar.org/CorpusID:268364256}
\BIBentrySTDinterwordspacing

\bibitem{Lu2023HumanTOMATOTW}
\BIBentryALTinterwordspacing
S.~Lu, L.-H. Chen, A.~Zeng, J.~de~Lin, R.~Zhang, L.~Zhang, and H.~yeung Shum, ``Humantomato: Text-aligned whole-body motion generation,'' \emph{ArXiv}, vol. abs/2310.12978, 2023. [Online]. Available: \url{https://api.semanticscholar.org/CorpusID:264306297}
\BIBentrySTDinterwordspacing

\bibitem{Liang2023OMGTO}
\BIBentryALTinterwordspacing
H.~Liang, J.~Bao, R.~Zhang, S.~Ren, Y.~Xu, S.~Yang, X.~Chen, J.~Yu, and L.~Xu, ``Omg: Towards open-vocabulary motion generation via mixture of controllers,'' \emph{2024 IEEE/CVF Conference on Computer Vision and Pattern Recognition (CVPR)}, pp. 482--493, 2023. [Online]. Available: \url{https://api.semanticscholar.org/CorpusID:266210422}
\BIBentrySTDinterwordspacing

\bibitem{makoviychuk2021isaac}
V.~Makoviychuk, L.~Wawrzyniak, Y.~Guo, M.~Lu, K.~Storey, M.~Macklin, D.~Hoeller, N.~Rudin, A.~Allshire, A.~Handa \emph{et~al.}, ``Isaac gym: High performance gpu-based physics simulation for robot learning,'' \emph{arXiv preprint arXiv:2108.10470}, 2021.

\bibitem{todorov2012mujoco}
E.~Todorov, T.~Erez, and Y.~Tassa, ``Mujoco: A physics engine for model-based control,'' in \emph{2012 IEEE/RSJ international conference on intelligent robots and systems}.\hskip 1em plus 0.5em minus 0.4em\relax IEEE, 2012, pp. 5026--5033.

\bibitem{Ghosh2023REMOS3M}
\BIBentryALTinterwordspacing
A.~Ghosh, R.~Dabral, V.~Golyanik, C.~Theobalt, and P.~Slusallek, ``Remos: 3d motion-conditioned reaction synthesis for two-person interactions,'' in \emph{European Conference on Computer Vision}, 2023. [Online]. Available: \url{https://api.semanticscholar.org/CorpusID:265466318}
\BIBentrySTDinterwordspacing

\bibitem{liang2024intergen}
H.~Liang, W.~Zhang, W.~Li, J.~Yu, and L.~Xu, ``Intergen: Diffusion-based multi-human motion generation under complex interactions,'' \emph{International Journal of Computer Vision}, pp. 1--21, 2024.

\bibitem{xu2024interx}
L.~Xu, X.~Lv, Y.~Yan, X.~Jin, S.~Wu, C.~Xu, Y.~Liu, Y.~Zhou, F.~Rao, X.~Sheng \emph{et~al.}, ``Inter-x: Towards versatile human-human interaction analysis,'' in \emph{Proceedings of the IEEE/CVF Conference on Computer Vision and Pattern Recognition}, 2024, pp. 22\,260--22\,271.

\bibitem{ruiz2024in2in}
P.~Ruiz-Ponce, G.~Barquero, C.~Palmero, S.~Escalera, and J.~Garc{\'\i}a-Rodr{\'\i}guez, ``in2in: Leveraging individual information to generate human interactions,'' in \emph{Proceedings of the IEEE/CVF Conference on Computer Vision and Pattern Recognition}, 2024, pp. 1941--1951.

\bibitem{Javed2024InterMask3H}
\BIBentryALTinterwordspacing
M.~G. Javed, C.~Guo, L.~Cheng, and X.~Li, ``Intermask: 3d human interaction generation via collaborative masked modelling,'' \emph{ArXiv}, vol. abs/2410.10010, 2024. [Online]. Available: \url{https://api.semanticscholar.org/CorpusID:273346671}
\BIBentrySTDinterwordspacing

\bibitem{Li2024InterDanceReactive3D}
\BIBentryALTinterwordspacing
R.~Li, Y.-Z. Zhang, Y.~Zhang, Y.~Zhang, M.~Su, J.~Guo, Z.~Liu, Y.~Liu, and X.~Li, ``Interdance:reactive 3d dance generation with realistic duet interactions,'' \emph{ArXiv}, vol. abs/2412.16982, 2024. [Online]. Available: \url{https://api.semanticscholar.org/CorpusID:274982678}
\BIBentrySTDinterwordspacing

\bibitem{Huang2024InterActCA}
\BIBentryALTinterwordspacing
Y.~Huang, L.~Ho, D.~Qin, M.~Shi, and T.~Komura, ``Interact: Capture and modelling of realistic, expressive and interactive activities between two persons in daily scenarios,'' \emph{ArXiv}, vol. abs/2405.11690, 2024. [Online]. Available: \url{https://api.semanticscholar.org/CorpusID:269921862}
\BIBentrySTDinterwordspacing

\bibitem{Xu2024ReGenNetTH}
\BIBentryALTinterwordspacing
L.~Xu, Y.~Zhou, Y.~Yan, X.~Jin, W.~Zhu, F.~Rao, X.~Yang, and W.~Zeng, ``Regennet: Towards human action-reaction synthesis,'' \emph{2024 IEEE/CVF Conference on Computer Vision and Pattern Recognition (CVPR)}, pp. 1759--1769, 2024. [Online]. Available: \url{https://api.semanticscholar.org/CorpusID:268531453}
\BIBentrySTDinterwordspacing

\bibitem{Tanaka2023RoleawareIG}
\BIBentryALTinterwordspacing
M.~Tanaka and K.~Fujiwara, ``Role-aware interaction generation from textual description,'' \emph{2023 IEEE/CVF International Conference on Computer Vision (ICCV)}, pp. 15\,953--15\,963, 2023. [Online]. Available: \url{https://api.semanticscholar.org/CorpusID:267024730}
\BIBentrySTDinterwordspacing

\bibitem{guo2022generating}
C.~Guo, S.~Zou, X.~Zuo, S.~Wang, W.~Ji, X.~Li, and L.~Cheng, ``Generating diverse and natural 3d human motions from text,'' in \emph{Proceedings of the IEEE/CVF Conference on Computer Vision and Pattern Recognition}, 2022, pp. 5152--5161.

\bibitem{huang2020dance}
R.~Huang, H.~Hu, W.~Wu, K.~Sawada, M.~Zhang, and D.~Jiang, ``Dance revolution: Long-term dance generation with music via curriculum learning,'' \emph{arXiv preprint arXiv:2006.06119}, 2020.

\bibitem{ghosh2021synthesis}
A.~Ghosh, N.~Cheema, C.~Oguz, C.~Theobalt, and P.~Slusallek, ``Synthesis of compositional animations from textual descriptions,'' in \emph{Proceedings of the IEEE/CVF international conference on computer vision}, 2021, pp. 1396--1406.

\bibitem{plappert2018learning}
M.~Plappert, C.~Mandery, and T.~Asfour, ``Learning a bidirectional mapping between human whole-body motion and natural language using deep recurrent neural networks,'' \emph{Robotics and Autonomous Systems}, vol. 109, pp. 13--26, 2018.

\bibitem{goodfellow2014gan}
I.~Goodfellow, J.~Pouget-Abadie, M.~Mirza, B.~Xu, D.~Warde-Farley, S.~Ozair, A.~Courville, and Y.~Bengio, ``Generative adversarial nets,'' \emph{Advances in neural information processing systems}, vol.~27, 2014.

\bibitem{Lopez2020vae}
\BIBentryALTinterwordspacing
R.~Lopez, P.~Boyeau, N.~Yosef, M.~I. Jordan, and J.~Regier, ``Auto-encoding variational bayes,'' 2020. [Online]. Available: \url{https://api.semanticscholar.org/CorpusID:211146177}
\BIBentrySTDinterwordspacing

\bibitem{Ho2020diffusion}
\BIBentryALTinterwordspacing
J.~Ho, A.~Jain, and P.~Abbeel, ``Denoising diffusion probabilistic models,'' \emph{ArXiv}, vol. abs/2006.11239, 2020. [Online]. Available: \url{https://api.semanticscholar.org/CorpusID:219955663}
\BIBentrySTDinterwordspacing

\bibitem{petrovich2022temos}
M.~Petrovich, M.~J. Black, and G.~Varol, ``Temos: Generating diverse human motions from textual descriptions,'' in \emph{European Conference on Computer Vision}.\hskip 1em plus 0.5em minus 0.4em\relax Springer, 2022, pp. 480--497.

\bibitem{chen2023mld}
X.~Chen, B.~Jiang, W.~Liu, Z.~Huang, B.~Fu, T.~Chen, and G.~Yu, ``Executing your commands via motion diffusion in latent space,'' in \emph{Proceedings of the IEEE/CVF conference on computer vision and pattern recognition}, 2023, pp. 18\,000--18\,010.

\bibitem{guo2024momask}
C.~Guo, Y.~Mu, M.~G. Javed, S.~Wang, and L.~Cheng, ``Momask: Generative masked modeling of 3d human motions,'' in \emph{Proceedings of the IEEE/CVF Conference on Computer Vision and Pattern Recognition}, 2024, pp. 1900--1910.

\bibitem{Lai2022RVQ}
\BIBentryALTinterwordspacing
C.-H. Lai, D.~Zou, and G.~Lerman, ``Robust vector quantized-variational autoencoder,'' \emph{ArXiv}, vol. abs/2202.01987, 2022. [Online]. Available: \url{https://api.semanticscholar.org/CorpusID:246608006}
\BIBentrySTDinterwordspacing

\bibitem{yuan2023physdiff}
Y.~Yuan, J.~Song, U.~Iqbal, A.~Vahdat, and J.~Kautz, ``Physdiff: Physics-guided human motion diffusion model,'' in \emph{Proceedings of the IEEE/CVF international conference on computer vision}, 2023, pp. 16\,010--16\,021.

\bibitem{wan2023diffusionphase}
W.~Wan, Y.~Huang, S.~Wu, T.~Komura, W.~Wang, D.~Jayaraman, and L.~Liu, ``Diffusionphase: Motion diffusion in frequency domain,'' \emph{arXiv preprint arXiv:2312.04036}, 2023.

\bibitem{cai2024digital}
Z.~Cai, J.~Jiang, Z.~Qing, X.~Guo, M.~Zhang, Z.~Lin, H.~Mei, C.~Wei, R.~Wang, W.~Yin \emph{et~al.}, ``Digital life project: Autonomous 3d characters with social intelligence,'' in \emph{Proceedings of the IEEE/CVF Conference on Computer Vision and Pattern Recognition}, 2024, pp. 582--592.

\bibitem{Chen2024SitcomCrafterAP}
\BIBentryALTinterwordspacing
J.~Chen, P.~Hu, X.~Chang, Z.~Shi, M.~C. Kampffmeyer, and X.~Liang, ``Sitcom-crafter: A plot-driven human motion generation system in 3d scenes,'' \emph{ArXiv}, vol. abs/2410.10790, 2024. [Online]. Available: \url{https://api.semanticscholar.org/CorpusID:273345513}
\BIBentrySTDinterwordspacing

\bibitem{mahmood2019amass}
N.~Mahmood, N.~Ghorbani, N.~F. Troje, G.~Pons-Moll, and M.~J. Black, ``Amass: Archive of motion capture as surface shapes,'' in \emph{Proceedings of the IEEE/CVF international conference on computer vision}, 2019, pp. 5442--5451.

\bibitem{yuan2020rfc}
Y.~Yuan and K.~Kitani, ``Residual force control for agile human behavior imitation and extended motion synthesis,'' \emph{Advances in Neural Information Processing Systems}, vol.~33, pp. 21\,763--21\,774, 2020.

\bibitem{peng2022ase}
X.~B. Peng, Y.~Guo, L.~Halper, S.~Levine, and S.~Fidler, ``Ase: Large-scale reusable adversarial skill embeddings for physically simulated characters,'' \emph{ACM Transactions On Graphics (TOG)}, vol.~41, no.~4, pp. 1--17, 2022.

\bibitem{peng2021amp}
X.~B. Peng, Z.~Ma, P.~Abbeel, S.~Levine, and A.~Kanazawa, ``Amp: Adversarial motion priors for stylized physics-based character control,'' \emph{ACM Transactions on Graphics (ToG)}, vol.~40, no.~4, pp. 1--20, 2021.

\bibitem{luo2023uhc}
Z.~Luo, J.~Cao, J.~Merel, A.~Winkler, J.~Huang, K.~Kitani, and W.~Xu, ``Universal humanoid motion representations for physics-based control,'' \emph{arXiv preprint arXiv:2310.04582}, 2023.

\bibitem{luo2023phc}
Z.~Luo, J.~Cao, K.~Kitani, W.~Xu \emph{et~al.}, ``Perpetual humanoid control for real-time simulated avatars,'' in \emph{Proceedings of the IEEE/CVF International Conference on Computer Vision}, 2023, pp. 10\,895--10\,904.

\bibitem{wang2023physhoi}
Y.~Wang, J.~Lin, A.~Zeng, Z.~Luo, J.~Zhang, and L.~Zhang, ``Physhoi: Physics-based imitation of dynamic human-object interaction,'' \emph{arXiv preprint arXiv:2312.04393}, 2023.

\bibitem{cui2024anyskill}
J.~Cui, T.~Liu, N.~Liu, Y.~Yang, Y.~Zhu, and S.~Huang, ``Anyskill: Learning open-vocabulary physical skill for interactive agents,'' in \emph{Proceedings of the IEEE/CVF Conference on Computer Vision and Pattern Recognition}, 2024, pp. 852--862.

\bibitem{yuan2023tennis}
Y.~YUAN and V.~Makoviychuk, ``Learning physically simulated tennis skills from broadcast videos,'' 2023.

\bibitem{yuan2021simpoe}
Y.~Yuan, S.-E. Wei, T.~Simon, K.~Kitani, and J.~Saragih, ``Simpoe: Simulated character control for 3d human pose estimation,'' in \emph{Proceedings of the IEEE/CVF conference on computer vision and pattern recognition}, 2021, pp. 7159--7169.

\bibitem{luo2022embodiedscene}
Z.~Luo, S.~Iwase, Y.~Yuan, and K.~Kitani, ``Embodied scene-aware human pose estimation,'' \emph{Advances in Neural Information Processing Systems}, vol.~35, pp. 6815--6828, 2022.

\bibitem{ugrinovic2024multiphys}
N.~Ugrinovic, B.~Pan, G.~Pavlakos, D.~Paschalidou, B.~Shen, J.~Sanchez-Riera, F.~Moreno-Noguer, and L.~Guibas, ``Multiphys: Multi-person physics-aware 3d motion estimation,'' in \emph{Proceedings of the IEEE/CVF Conference on Computer Vision and Pattern Recognition}, 2024, pp. 2331--2340.

\bibitem{yuan2019ego}
Y.~Yuan and K.~Kitani, ``Ego-pose estimation and forecasting as real-time pd control,'' in \emph{Proceedings of the IEEE/CVF International Conference on Computer Vision}, 2019, pp. 10\,082--10\,092.

\bibitem{liu2023emage}
H.~Liu, Z.~Zhu, G.~Becherini, Y.~Peng, M.~Su, Y.~Zhou, X.~Zhe, N.~Iwamoto, B.~Zheng, and M.~J. Black, ``Emage: Towards unified holistic co-speech gesture generation via masked audio gesture modeling,'' \emph{arXiv e-prints}, pp. arXiv--2401, 2023.

\bibitem{yi2023talkshow}
H.~Yi, H.~Liang, Y.~Liu, Q.~Cao, Y.~Wen, T.~Bolkart, D.~Tao, and M.~J. Black, ``Generating holistic 3d human motion from speech,'' in \emph{Proceedings of the IEEE/CVF Conference on Computer Vision and Pattern Recognition}, 2023, pp. 469--480.

\bibitem{qi2024cocogesture}
X.~Qi, H.~Zhang, Y.~Wang, J.~Pan, C.~Liu, P.~Li, X.~Chi, M.~Li, Q.~Zhang, W.~Xue \emph{et~al.}, ``Cocogesture: Toward coherent co-speech 3d gesture generation in the wild,'' \emph{arXiv preprint arXiv:2405.16874}, 2024.

\bibitem{loper2015smpl}
M.~Loper, N.~Mahmood, J.~Romero, G.~Pons-Moll, and M.~J. Black, ``Smpl: A skinned multi-person linear model,'' \emph{ACM transactions on graphics (TOG)}, vol.~34, no.~6, pp. 1--16, 2015.

\bibitem{smplx}
G.~Pavlakos, V.~Choutas, N.~Ghorbani, T.~Bolkart, A.~A. Osman, D.~Tzionas, and M.~J. Black, ``Expressive body capture: 3d hands, face, and body from a single image,'' in \emph{Proceedings of the IEEE/CVF conference on computer vision and pattern recognition}, 2019, pp. 10\,975--10\,985.

\bibitem{peng2018deepmimic}
X.~B. Peng, P.~Abbeel, S.~Levine, and M.~Van~de Panne, ``Deepmimic: Example-guided deep reinforcement learning of physics-based character skills,'' \emph{ACM Transactions On Graphics (TOG)}, vol.~37, no.~4, pp. 1--14, 2018.

\bibitem{fu2023deepwholebody}
Z.~Fu, X.~Cheng, and D.~Pathak, ``Deep whole-body control: learning a unified policy for manipulation and locomotion,'' in \emph{Conference on Robot Learning}.\hskip 1em plus 0.5em minus 0.4em\relax PMLR, 2023, pp. 138--149.

\bibitem{bogo2016keep}
F.~Bogo, A.~Kanazawa, C.~Lassner, P.~Gehler, J.~Romero, and M.~J. Black, ``Keep it smpl: Automatic estimation of 3d human pose and shape from a single image,'' in \emph{Computer Vision--ECCV 2016: 14th European Conference, Amsterdam, The Netherlands, October 11-14, 2016, Proceedings, Part V 14}.\hskip 1em plus 0.5em minus 0.4em\relax Springer, 2016, pp. 561--578.

\bibitem{radford2021clip}
A.~Radford, J.~W. Kim, C.~Hallacy, A.~Ramesh, G.~Goh, S.~Agarwal, G.~Sastry, A.~Askell, P.~Mishkin, J.~Clark \emph{et~al.}, ``Learning transferable visual models from natural language supervision,'' in \emph{International conference on machine learning}.\hskip 1em plus 0.5em minus 0.4em\relax PMLR, 2021, pp. 8748--8763.

\bibitem{wang2023refit}
Y.~Wang and K.~Daniilidis, ``Refit: Recurrent fitting network for 3d human recovery,'' in \emph{Proceedings of the IEEE/CVF International Conference on Computer Vision}, 2023, pp. 14\,644--14\,654.

\bibitem{yao2023whmr}
W.~Yao, H.~Zhang, Y.~Sun, and J.~Tang, ``W-hmr: Human mesh recovery in world space with weak-supervised camera calibration and orientation correction,'' \emph{arXiv preprint arXiv:2311.17460}, 2023.

\bibitem{li2021ai}
R.~Li, S.~Yang, D.~A. Ross, and A.~Kanazawa, ``Ai choreographer: Music conditioned 3d dance generation with aist++,'' in \emph{Proceedings of the IEEE/CVF International Conference on Computer Vision}, 2021, pp. 13\,401--13\,412.

\bibitem{tseng2023edge}
J.~Tseng, R.~Castellon, and K.~Liu, ``Edge: Editable dance generation from music,'' in \emph{Proceedings of the IEEE/CVF Conference on Computer Vision and Pattern Recognition}, 2023, pp. 448--458.

\bibitem{meng2024rethinking}
Z.~Meng, Y.~Xie, X.~Peng, Z.~Han, and H.~Jiang, ``Rethinking diffusion for text-driven human motion generation,'' \emph{arXiv preprint arXiv:2411.16575}, 2024.

\bibitem{zheng2024beat}
C.~Zheng, J.~Qin, and S.~He, ``Beat-it: Beat-synchronized multi-condition 3d dance generation,'' 2024.

\bibitem{guo2020action2motion}
C.~Guo, X.~Zuo, S.~Wang, S.~Zou, Q.~Sun, A.~Deng, M.~Gong, and L.~Cheng, ``Action2motion: Conditioned generation of 3d human motions,'' in \emph{Proceedings of the 28th ACM International Conference on Multimedia}, 2020, pp. 2021--2029.

\bibitem{lee2019dancing}
H.-Y. Lee, X.~Yang, M.-Y. Liu, T.-C. Wang, Y.-D. Lu, M.-H. Yang, and J.~Kautz, ``Dancing to music,'' \emph{Advances in neural information processing systems}, vol.~32, 2019.

\end{thebibliography}












\begin{IEEEbiography}[{\includegraphics[width=1in,height=1.25in,clip,keepaspectratio]{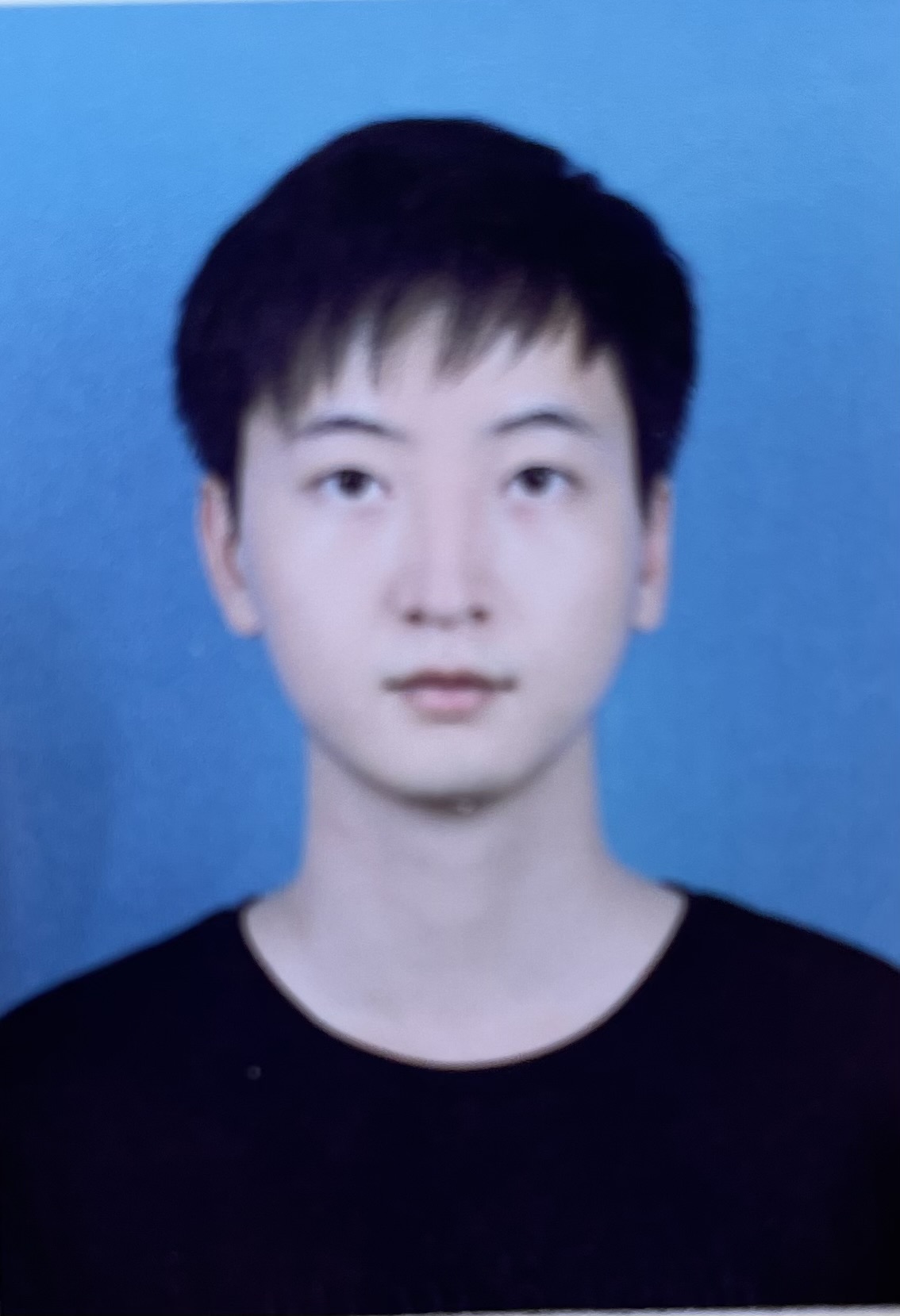}}]{Wei Yao} received the B.E. degree from the University of South China, Hengyang, China, in 2021. He is now a Ph.D. student in the School of Computer Science and Engineering at Nanjing University of Science and Technology, Nanjing, China. His research interests include computer vision, motion capture and embodied intelligence
\end{IEEEbiography}

\begin{IEEEbiography}[{\includegraphics[width=1in,height=1.25in,clip,keepaspectratio]{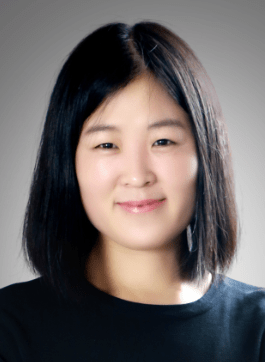}}]{Yunlian Sun} received the ME degree in computer science and technology from the Harbin Institute of Technology, China, in 2010 and the Ph.D. degree in ingegneria elettronica, informatica e delle telecomunicazioni from the University of Bologna, Italy, in 2014. After the Ph.D study, she worked as a postdoctoral researcher at National Laboratory of Pattern Recognition (NLPR), Institute of Automation, Chinese Academy of Sciences. She is currently an Associate Professor at the School of Computer Science and Engineering, Nanjing University of Science and Technology, China. Her research interests include biometrics, pattern recognition, and computer vision. 
\end{IEEEbiography}

\begin{IEEEbiography}[{\includegraphics[width=1in,height=1.25in,clip,keepaspectratio]{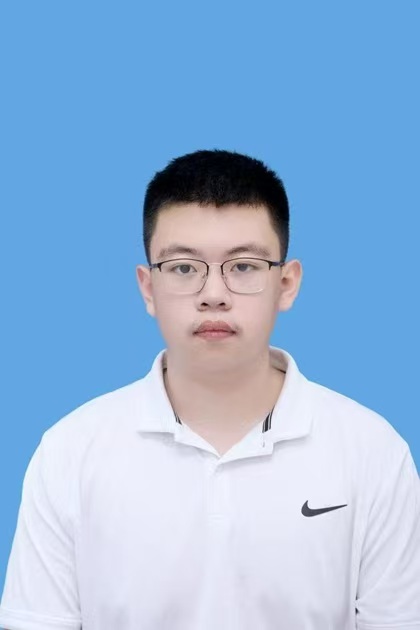}}]{Chang Liu} is a junior undergraduate student at Beijing Normal University. His research interests include 3D reconstruction,depth estimation,and digital human generation.He is currently working on digital human research as an intern in the Department of Automation,Tsinghua University and the School of Artificial Intelligence at Beijing Normal University. 
\end{IEEEbiography}

\begin{IEEEbiography}[{\includegraphics[width=1in,height=1.25in,clip,keepaspectratio]{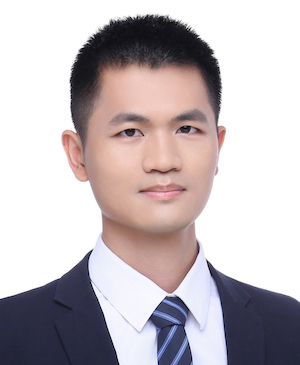}}]{Hongwen Zhang} received the B.E. degree from the South China University of Technology, Guangzhou, China, in 2015, and the Ph.D. degree from the Institute of Automation, Chinese Academy of Sciences, Beijing, China, in 2021, respectively. He has been working as a Post-Doctoral Researcher at Tsinghua University and is currently an Associate Professor at the School of Artificial Intelligence, Beijing Normal University. His research interests include computer vision, computer graphics, and their applications in 3D human modeling.
\end{IEEEbiography}

\begin{IEEEbiography}[{\includegraphics[width=1in,height=1.25in,clip,keepaspectratio]{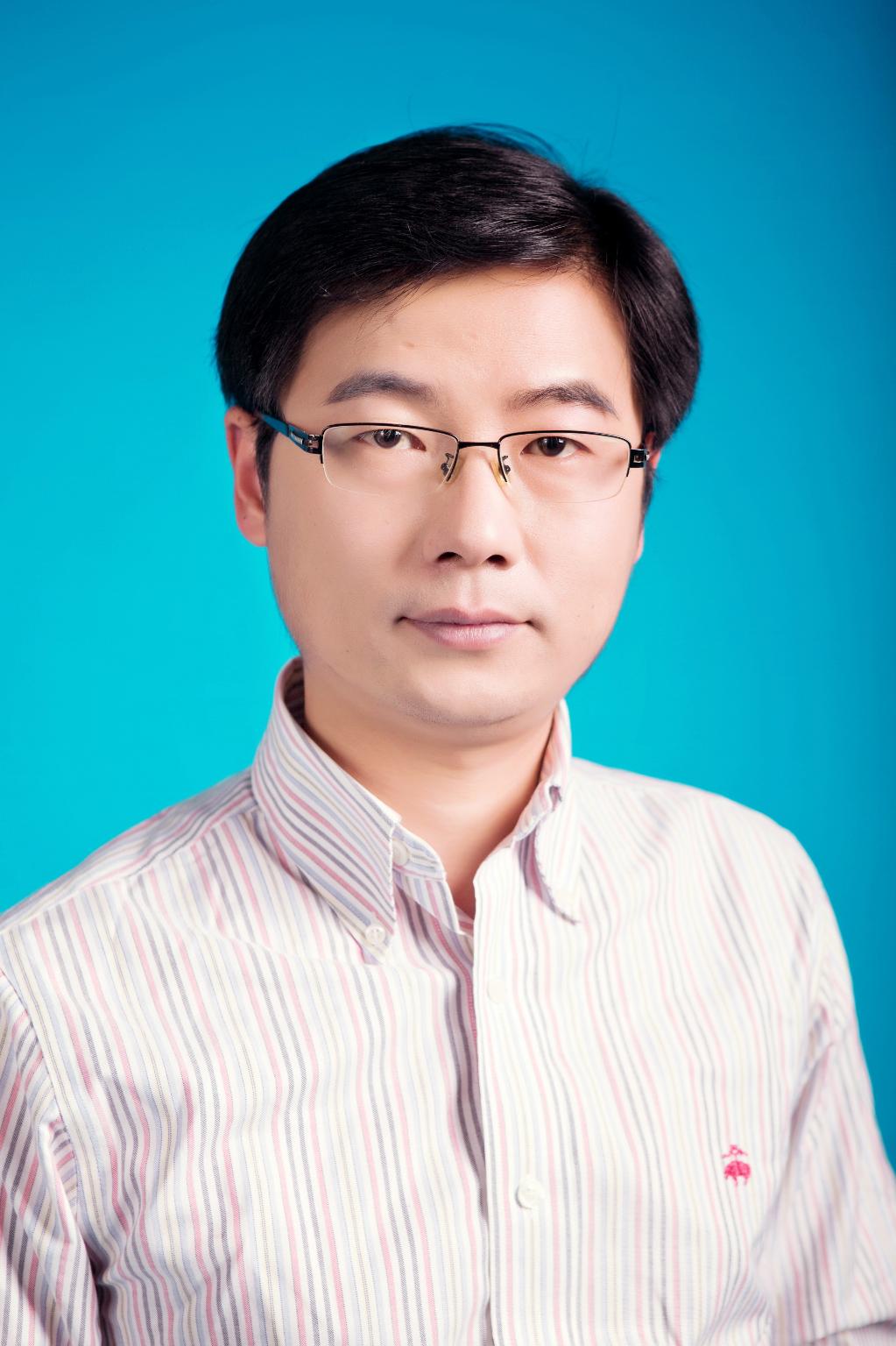}}]{Jinhui Tang} (Senior Member, IEEE) received the B.E. and Ph.D. degrees from the University of Science and Technology of China, Hefei, China, in 2003 and 2008, respectively. He is currently a Professor with the Nanjing University of Science and Technology, Nanjing, China. He has authored more than 200 articles in toptier journals and conferences. His research interests include multimedia analysis and computer vision. Dr.Tang was a recipient of the Best Paper Awards in ACM MM 2007 and ACM MM Asia 2020, the Best Paper Runner-Up in ACM MM 2015. He has served as an Associate Editor for the IEEE TNNLS, IEEE TKDE, IEEE TMM, and IEEE TCSVT. He is a Fellow of IAPR. 
\end{IEEEbiography}

\end{document}